\documentclass[10pt,twocolumn,letterpaper]{article}

\usepackage{iccv}
\usepackage{times}
\usepackage{epsfig}
\usepackage{graphicx}
\usepackage{amsmath}
\usepackage{amssymb}

\usepackage{graphicx}
\usepackage{booktabs}
\usepackage{bbold}
\usepackage{array}
\usepackage{graphicx, amsmath, amssymb, caption, subcaption, multirow, overpic, textpos}
\usepackage[table]{xcolor}
\usepackage{colortbl}
\usepackage[linesnumbered,ruled,vlined]{algorithm2e}
\SetKwInOut{Parameter}{Parameters}
\usepackage{listings}
\usepackage{tikz}

\newcommand\blfootnote[1]{%
  \begingroup
  \renewcommand\thefootnote{}\footnote{#1}%
  \addtocounter{footnote}{-1}%
  \endgroup
}
\DeclareMathOperator*{\argmax}{argmax}

\definecolor{darkred}{rgb}{0.8,0.02,0.02}
\definecolor{Gray}{gray}{0.9}

\definecolor{MyGreen}{rgb}{0,0.6,0.3}
\definecolor{blackpink}{rgb}{0.6,0,0.6}
\definecolor{blue}{RGB}{0, 0, 200}
\definecolor{red}{RGB}{255, 0, 0}
\definecolor{black}{RGB}{0, 0, 0}

\definecolor{pink}{RGB}{247, 159, 166}
\definecolor{red}{RGB}{192, 0, 0}

\def\eg{\textit{e.g., }}
\def\ie{\textit{i.e., }}
\newcommand\edit[1]{\textcolor{red}{#1}}

\usepackage{xparse}
\NewDocumentCommand{\framecolorbox}{oommm}
 {% #1 = width (optional)
  \IfValueTF{#1}
   {%
    \IfValueTF{#2}
     {\fcolorbox{#3}{#4}{\makebox[#1][#2]{#5}}}
     {\fcolorbox{#3}{#4}{\makebox[#1]{#5}}}%
   }
   {\fcolorbox{#3}{#4}{#5}}%
 }

\usepackage[pagebackref=true,breaklinks=true,letterpaper=true,colorlinks,bookmarks=false]{hyperref}

\usepackage[
  separate-uncertainty = true,
  multi-part-units = repeat
]{siunitx}

\usepackage{makecell}
\iccvfinalcopy % *** Uncomment this line for the final submission

\def\iccvPaperID{****} % *** Enter the ICCV Paper ID here

\ificcvfinal\fi
\newcommand*{\affmark}[1][*]{\textsuperscript{#1}}

\begin{document}

%%%%%%%%% TITLE
\title{Towards Open-Set Test-Time Adaptation \\ Utilizing the Wisdom of Crowds in Entropy Minimization}

\author{
Jungsoo Lee\affmark[1,2]$^\text{*}$ \; Debasmit Das\affmark[1] \; Jaegul Choo\affmark[2] \; Sungha Choi\affmark[1]$^\dagger$ \; \vspace{0.2cm}\\
\affmark[1]Qualcomm AI Research$^\ddagger$ \; \affmark[2]KAIST\\
\texttt{\footnotesize\affmark[1]\{jungsool, debadas, sunghac\}@qti.qualcomm.com} \; \texttt{\footnotesize\affmark[2]\{bebeto, jchoo\}@kaist.ac.kr}\\
% \vspace*{-1.5cm}
}

\makeatletter
\g@addto@macro\@maketitle{
% \begin{strip}
    \centering
    \vspace{-0.9cm}
    \includegraphics[width=\textwidth]{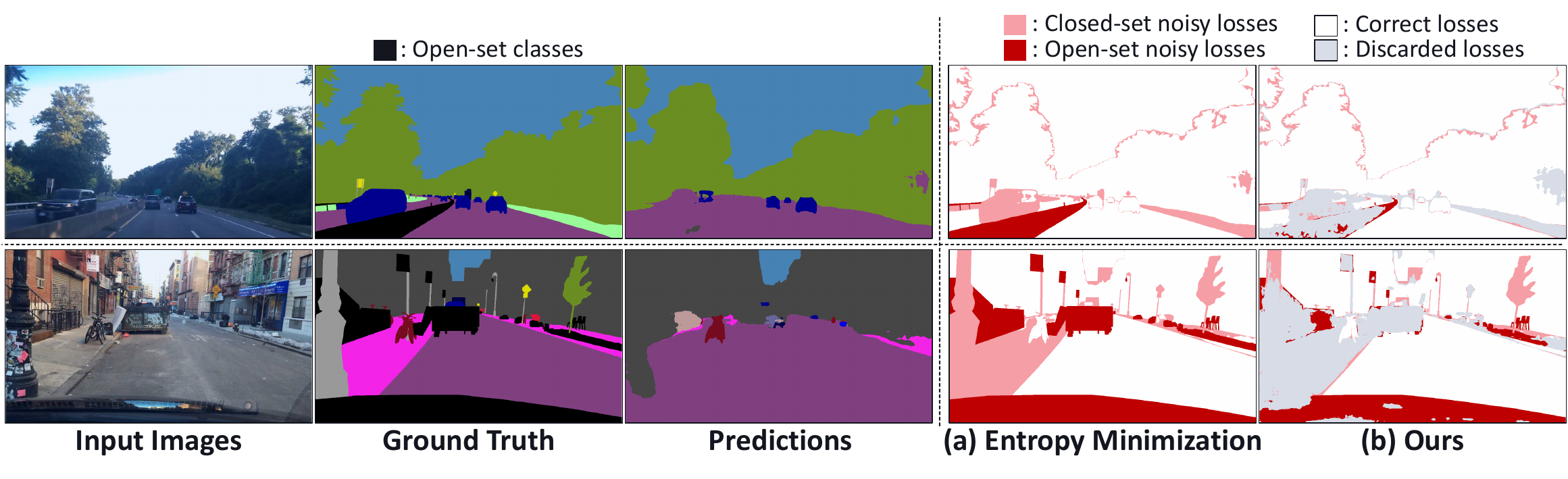}
    \vspace{-0.75cm}
    \captionof{figure}{Discarding noisy losses in test-time adaptation (TTA). Black pixels in the ground truth indicate the open-set classes in the test set (\ie BDD-100K~\cite{bdd100k}), which were not included in the train set (\ie Cityscapes~\cite{cityscapes}), such as the guardrails in the first row or the garbage truck in the second row. (a) Since TTA models generally use their predictions for the target outputs, they are prone to utilizing noisy losses from 1) wrong predictions (pink pixels \textcolor{pink}{$\blacksquare$}) and 2) open-set classes (red pixels \textcolor{red}{$\blacksquare$}). Performing TTA for a long term in such an environment degrades the performance of TTA models significantly. 
    (b) Our method effectively filters out such noisy losses, preventing performance degradation and alarming unexpected obstacles, which is crucial for safety-critical applications such as autonomous driving (see Supplementary).}
    \vspace{0.2cm}
    \label{fig:teaser}
}

\maketitle

\blfootnote{$^*$Work done during an internship at Qualcomm AI Research.}
\blfootnote{$^\dagger$ Corresponding author. \hspace{0.1cm} $^\ddagger$ Qualcomm AI Research is an initiative of Qualcomm Technologies, Inc.}

\ificcvfinal\thispagestyle{empty}\fi

%%%%%%%%% ABSTRACT
\vspace{-0.4cm}
\begin{abstract}
    \vspace{-0.3cm}
    Test-time adaptation (TTA) methods, which generally rely on the model's predictions (\eg entropy minimization) to adapt the source pretrained model to the unlabeled target domain, suffer from noisy signals originating from 1) incorrect or 2) open-set predictions.
    Long-term stable adaptation is hampered by such noisy signals, so training models without such error accumulation is crucial for practical TTA.
    To address these issues, including open-set TTA, we propose a \textit{simple yet effective} sample selection method inspired by the following crucial empirical finding.
    While entropy minimization compels the model to increase the probability of its predicted label (\ie confidence values), we found that noisy samples rather show decreased confidence values.
    To be more specific, entropy minimization attempts to raise the confidence values of an individual sample's prediction, but individual confidence values may rise or fall due to the influence of signals from numerous other predictions (\ie wisdom of crowds).
    Due to this fact, noisy signals misaligned with such `wisdom of crowds', generally found in the correct signals, fail to raise the individual confidence values of wrong samples, despite attempts to increase them.
    Based on such findings, we filter out the samples whose confidence values are lower in the adapted model than in the original model, as they are likely to be noisy. 
    Our method is widely applicable to existing TTA methods and improves their long-term adaptation performance in both image classification (\eg 49.4\% reduced error rates with TENT) and semantic segmentation (\eg 11.7\% gain in mIoU with TENT). 
\end{abstract}

%%%%%%%%% BODY TEXT
\vspace{-0.6cm}
\section{Introduction}
\vspace{-0.1cm}
\begin{figure*}
    \centering
    \includegraphics[width=0.97\textwidth]{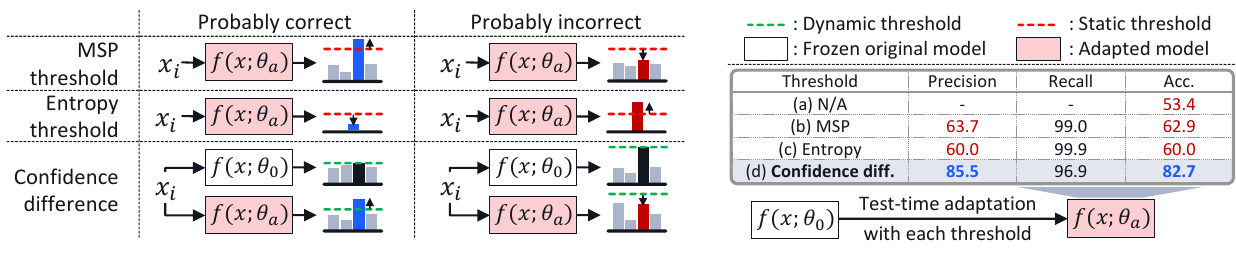}
    \vspace{-0.35cm}    
    \captionof{figure}{Utilizing confidence difference for selecting correct samples. Pseudo-labeling samples (\ie selecting correct samples) by using a fixed threshold does not guarantee a reasonable level of pseudo-labeling performance, which is demonstrated by the significantly low precision values. On the other hand, we maintain a reasonable level of both precision and recall by using the confidence difference between $\theta_o$ and $\theta_a$, improving the test-time adaptation performance overall.}
    \vspace{-0.4cm}
    \label{fig:motivation}
\end{figure*}
\vspace{-0.1cm}

% TTA task 
Despite the recent advancements of deep learning, models still show a significant performance degradation when confronted with large domain shifts (\eg changes of cities with different landscapes during autonomous driving)~\cite{robustnet, lee2022wildnet,lim2023ttn,pin-in-memory,transadapt}. 
Among various studies, test-time adaptation (TTA) is at the center of attention due to its practicality in not requiring 1) the source data during the adaptation stage and 2) ground truth labels of the target domain~\cite{tent}.

\begin{figure}[b]
    \centering
    \vspace{-0.6cm}
    \includegraphics[width=0.49\textwidth]{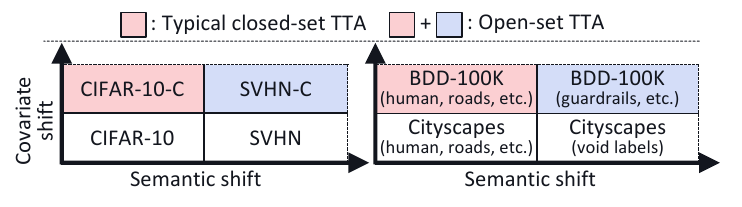}
    \vspace{-0.75cm}
    \caption{Description of open-set TTA. While previous studies assume covariate shifts (\ie Cityscapes to BDD-100K), they fail to address the semantic shifts (\ie guardrails only shown in BDD-100K). This paper addresses both closed-set and open-set test-time adaptation.}
    \label{fig:motivation_method_open}
    \vspace{-0.15cm}
\end{figure}

% TTA models widely utilize self-training strategy (\eg entropy minimization), which uses the predicted outputs as the target outputs in the loss functions~\cite{tent,swr,cotta,eata,conjugate}. 
TTA models widely utilize a self-training strategy (\eg entropy minimization), which uses the model's prediction as the target of the loss function~\cite{tent,swr,cotta,eata,conjugate,note,zhang2021memo, wildtta,delta,cafa}.
Since TTA models rely on their own predictions during the adaptation, they are inevitably prone to utilizing noisy signals.
In this paper, noisy signals indicate supervisions that originated from 1) incorrect or 2) open-set predictions. 
Fig.~\ref{fig:teaser} shows that performing adaptation with such noisy signals significantly degrades the TTA performance. 
Specifically, the pink pixels indicate the mispredicted pixels (\eg predicting sidewalks as roads in the second row), and the red ones are the predictions on open-set classes that were not included in the train set (\eg predicting guardrails and the garbage truck as roads in the first and second rows, respectively).
% Such an example clearly demonstrates that TTA in real-world applications needs to address such open-set classes since mispredicting guardrails as roads may cause serious accidents if occurred during autonomous driving.
% Adaptation to test samples of unknown classes (\ie open-set test-time adaptation) has not been explored in existing TTA studies despite its importance and practicality.
Such an example clearly demonstrates that TTA in real-world applications needs to address such open-set classes since mispredicting guardrails as roads may cause serious accidents during autonomous driving. 
However, as shown in Fig.~\ref{fig:motivation_method_open}, previous studies focused on TTA with covariate shifts (\ie domain shifts) only and did not address TTA that also includes semantic shifts (\ie including open-set classes).
Regarding its significance and practicality, adaptation with unknown classes included (\ie open-set TTA) should be also addressed.
% Despite its significance and practicality, adaptation to test samples of unknown classes (\ie open-set TTA) has not been studied previously.
% Therefore, we need an approach that effectively discards noisy signals to enable stable adaptation. 

% Findings
Fig.~\ref{fig:motivation} shows our empirical analysis that discloses an important finding to address such an issue. 
While entropy minimization enforces the model to increase the probability value of its predicted label (\ie confidence values), we found that it often fails to increase them on the wrong samples. 
% While previous studies~\cite{cotta, eata} resorted to finding an adequate confidence value or loss value for selecting correct samples, utilizing such a \textit{static} threshold shows limited performance.
While previous studies~\cite{cotta, eata} resorted to finding an adequate confidence value or loss value to prevent error accumulation, the process of determining it is cumbersome, and utilizing such a \textit{static} threshold shows limited performance.
% for selecting them. 
We train TENT~\cite{tent} with different thresholds for the analysis: (a) without thresholding, (b) selecting samples with confidence value higher or equal to 0.9\footnote{We used the best confidence value $p$ after grid search of $p \in \{0.5, 0.8, 0.9, 0.95, 0.99.\}$\vspace{-0.4cm}}, (c) selecting samples with loss values smaller than the entropy threshold proposed in EATA~\cite{eata}, and (d) selecting samples that achieve higher confidence value with the adaptation model $\theta_{a}$ compared to that with the original model $\theta_{o}$. 
As shown, using the confidence difference between $\theta_o$ and $\theta_a$ for selecting correct samples outperforms utilizing the static thresholds.
While b) and c) show significantly high recall values (\ie selecting actual correct samples well), it rather indicates that they simply select most of the samples and fail to filter out noisy samples considering the substantially low precision values (\ie low ratio of correct samples among the selected ones). 

% Fig.~\ref{fig:motivation_confidence_change} shows the confidence tendency between the adaptation model $\theta_a$ and the original model $\theta_o$. 
% The left figure shows the samples which $\theta_a$ achieves higher probability value on the predicted label of $\theta_o$. 
% We can observe that most of the samples are the correct ones in such a case. 
% On the other hand, the right figure shows that most of the samples which $\theta_a$ achieves lower probability value on the predicted label of $\theta_o$ are wrong samples. 
% In other words, correct samples show increased probability values on its originally predicted label while the wrong ones show decreased values.
The intuition behind using the confidence difference is as follows.
% Given a certain predicted label, the model learns the common pattern of correct samples and such signals reach to a consensus which are helpful to increase the confidence value on the class.
% Given a certain predicted label, the signals from correct samples reach to a consensus, which are helpful to increase the confidence value on the predicted class.
% On the other hand, the signals from the noisy samples misalign with such correct signals learned from the correct samples. 
% In other words, although the supervision from each individual sample enforces the model to increase the confidence level for each sample, they are nullified by the \textit{`wisdom of crowds'} learned from the majority of the correct samples.
Although entropy minimization enforces the model to increase the confidence value on the predicted label of an individual sample, the individual confidence value may rise or fall, influenced by the signals that originated from numerous other predictions (\ie wisdom of crowds).
To be more specific, the noisy signals that do not align with such `wisdom of crowds', commonly found in the correct signals, fail to raise the individual confidence scores of wrong samples, even with the supervision from entropy minimization to increase them.
By using such an observation, we select samples that achieve higher confidence value using $\theta_a$ compared to that using $\theta_o$. 
Since we reflect the knowledge state of the model on each individual sample, our selection is implicitly a dynamic thresholding strategy, which outperforms the previously-used static strategies. 
Our ~\textit{simple yet effective} sample selection method is widely applicable to existing TTA methods and improves their performances on both image classification and semantic segmentation.
% Based on such findings, this paper proposes a \textit{simple yet effective} sample selection method which filters out noisy samples by using the confidence difference between $\theta_o$ and $\theta_a$.
% Our method simply filters out samples achieving lower probability values on the originally predicted labels since the losses generated from such samples misalign with the wisdom of crowds found in correct samples. 

\begin{figure}[t]
    \centering
    \includegraphics[width=0.49\textwidth]{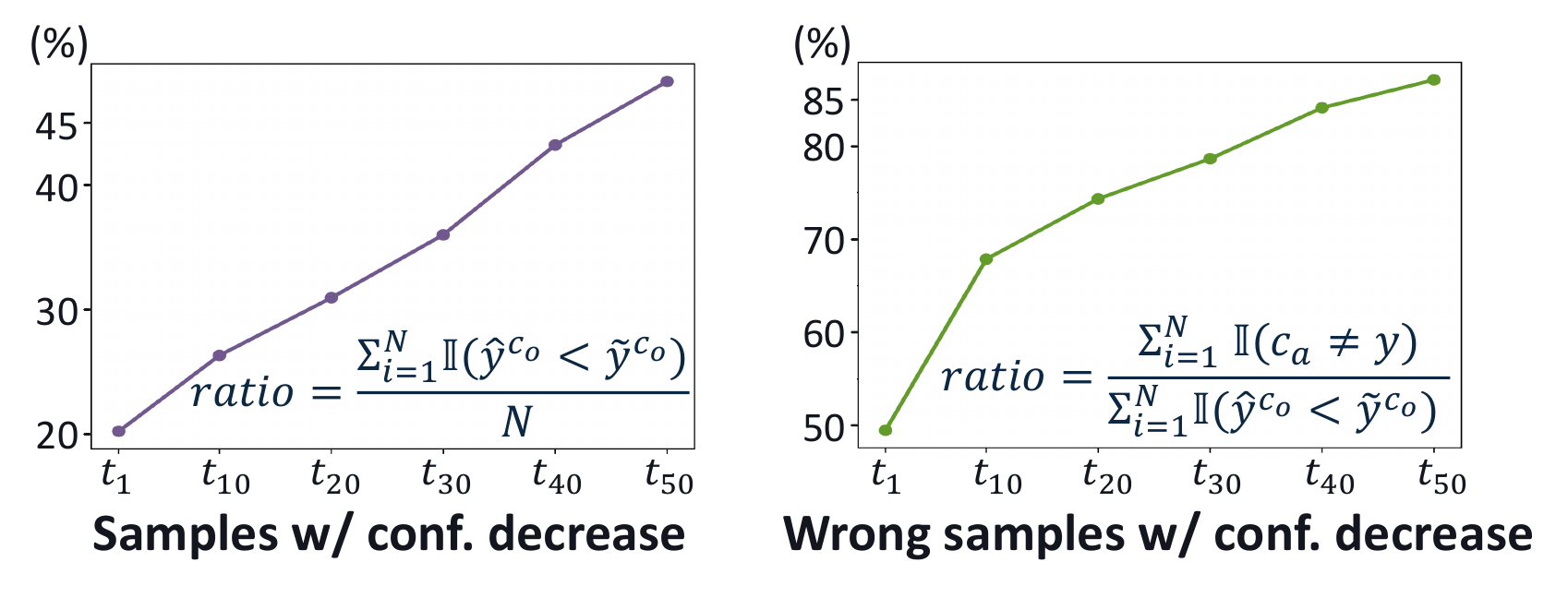}
    \vspace{-0.75cm}
    \caption{As adaptation proceeds, the number of samples with decreased confidence values increases (purple graph). Additionally, among those samples, the ratio of wrongly predicted samples also increases (green graph). $t_{i}$ indicates the $i^{th}$ round during the long-term adaptation.}
    \label{fig:motivation_method_npv}
    \vspace{-0.35cm}
\end{figure}

% Contributions
Our contributions are summarized as follows:
\begin{itemize}
    \vspace{-0.2cm}
    \item We propose a novel sample selection method that filters out noisy samples using the confidence difference between $\theta_{a}$ and $\theta_{o}$ based on the finding that noisy samples, both closed-set wrong samples, and open-set samples, generally show decreased confidence values on the originally predicted label.  
    \vspace{-0.2cm}
    % \item We propose the open-set test-time adaptation, an evaluation setting that conducts experiments including unknown classes with the same domain shift.
    % \item We propose the open-set test-time adaptation, adapting to a target domain including test samples of unknown classes, which has not been explored in existing TTA studies despite its importance and practicality.
    \item This is the first paper to address open-set test-time adaptation, adapting to a target domain including test samples of unknown classes, which has not been explored in existing TTA studies despite its importance and practicality.    
    \vspace{-0.2cm}
    \item Our proposed method can be applied to various test-time adaptation methods and improves their performances on both image classification using CIFAR-10/100-C and TinyImageNet-C (\eg 49.38\% reduced error rates with TENT in open-set TTA), and semantic segmentation (\eg 11.69\% gain in mIoU with TENT) using real-world datasets including BDD-100K and Mapillary.  
\end{itemize}

% outperforms the existing TTA methods on 1) image classification using CIFAR-10/100-C, and Tiny Imagenet-C and 2) semantic segmentation using test sets of BDD-100K~\cite{bdd100k}, Mapillary~\cite{mapillary}, GTAV~\cite{gta5}, SYNTHIA~\cite{synthia} on a model pretrained with Cityscapes~\cite{cityscapes}.

\vspace{-0.1cm}
\section{Wisdom of Crowds in Entropy Minimization}
\subsection{Problem Setup}
% Data x_i y_i no pair input
During the test-time adaptation, models adapt to a target domain with $N$ number of test samples in the test set $D_T$, $\{x_i, \}^{N}_{i=1} \in D_T$, without target labels provided.
Given a pretrained model $\theta_o$, we update $\theta_o$ to adapt to a novel target domain, where the adapted model is then defined as $\theta_a$. 
%%y -> tilde / hat 
% We define the softmax output of the test sample $x$ using the original model $\theta_o$ as $\Tilde{y}=f(x; \theta_o) \in \mathbb{R^{C}}$ with $C$ number of target classes, respectively.
For a test sample $x$, we define $\Tilde{y}=f(x; \theta_o) \in \mathbb{R}^{C}$ and $\hat{y}=f(x; \theta_a) \in \mathbb{R}^{C}$ as the softmax outputs of the original model $\theta_o$ and the adapted model $\theta_a$, respectively, where $C$ denotes the number of classes.
% With the predicted class of the original model $c_o=\argmax_c f(x; \theta_o)$, we define the probability value on the predicted label of $\theta_o$ as $\Tilde{y}^{c_o}$.
With the predicted class $c_o=\argmax_c f(x; \theta_o)$ of the original model, we define the probability value on the predicted label as confidence value $\Tilde{y}^{c_o}$.
% Similarly, for the adapted model $\theta_a$, the predicted class and the confidence value of the predicted label are defined as $c_a=\argmax_c f(x; \theta_a)$ and $\hat{y}^{c_a}$, respectively.
Similarly, the confidence value of the adapted model $\theta_a$ on the label $c_o$, predicted by the original model, is defined as $\hat{y}^{c_o}$.
% With the predicted label of the original model $y_{o}$, we define the probability value on the predicted label of $\theta_o$ as $f^{y_{o}}(x; \theta_o)$ and that of $\theta_a$ on $\Tilde{y}$ is defined as $f^{\Tilde{y}}(x; \theta_a)$.
%%% max y_o  
% $y^{c_o}_a$
The main objective of test-time adaptation is to correctly predict $c_a=\argmax_c f(x; \theta_a)$ using the adapted model, especially under large data distribution shifts.

\begin{figure}[t]
    \centering
    \includegraphics[width=0.49\textwidth]{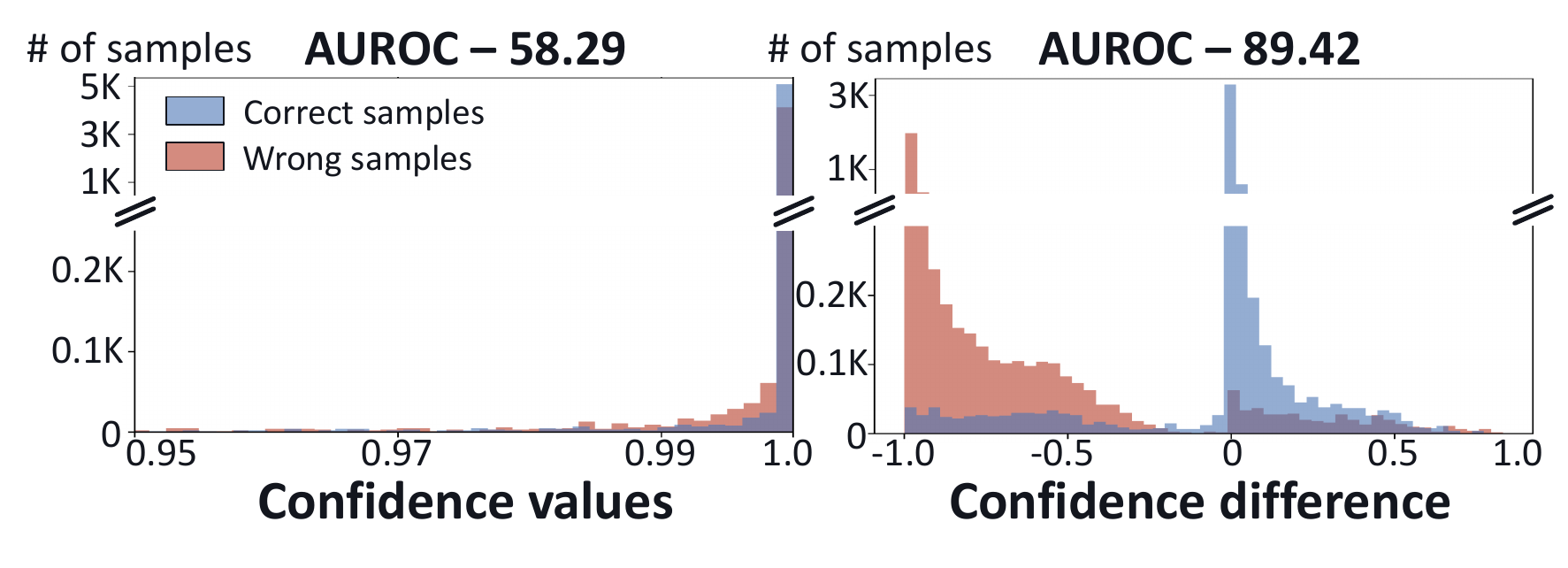}
    \vspace{-0.75cm}
    \caption{Utilizing the confidence difference distinguishes between the correct samples (blue) and the wrong samples (red) better (AUROC of 89.42) than using the confidence values (AUROC of 58.29). We used the same model (\ie TENT~\cite{tent} adapted for 50 rounds) for the visualization.}
    \label{fig:motivation_method_histogram}
    \vspace{-0.35cm}
\end{figure}

% \subsection{Wisdom of Crowds in Confidence Difference}
% \subsection{\sh{Wisdom of Crowds in Entropy Minimization}}
\subsection{Motivation}
\vspace{-0.05cm}

\begin{figure*}[t]
    \centering
    % \vspace{-1cm}
    \includegraphics[width=0.95\textwidth]{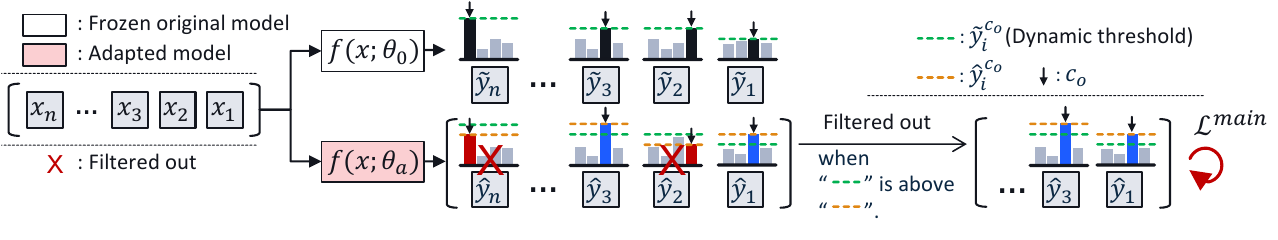}
    \vspace{-0.5cm}
    \caption{Overall procedure of our sample selection. We forward the mini-batch of $n$ test images, $\{x_i\}^{n}_{i=1}$, to the original model $\theta_o$ and the adaptation model $\theta_a$. Then, we compare the probability values $\hat{y}^{c_o}$ and $\Tilde{y}^{c_o}$ and select the samples achieving ${\hat{y}^{c_o} \geq \Tilde{y}^{c_o}}$. Finally, we apply the entropy minimization only to the selected samples.}
    \vspace{-0.25cm}
    \label{fig:method}
\end{figure*}

\begin{figure}[b]
    \centering
    \vspace{-0.3cm}
    \includegraphics[width=0.49\textwidth]{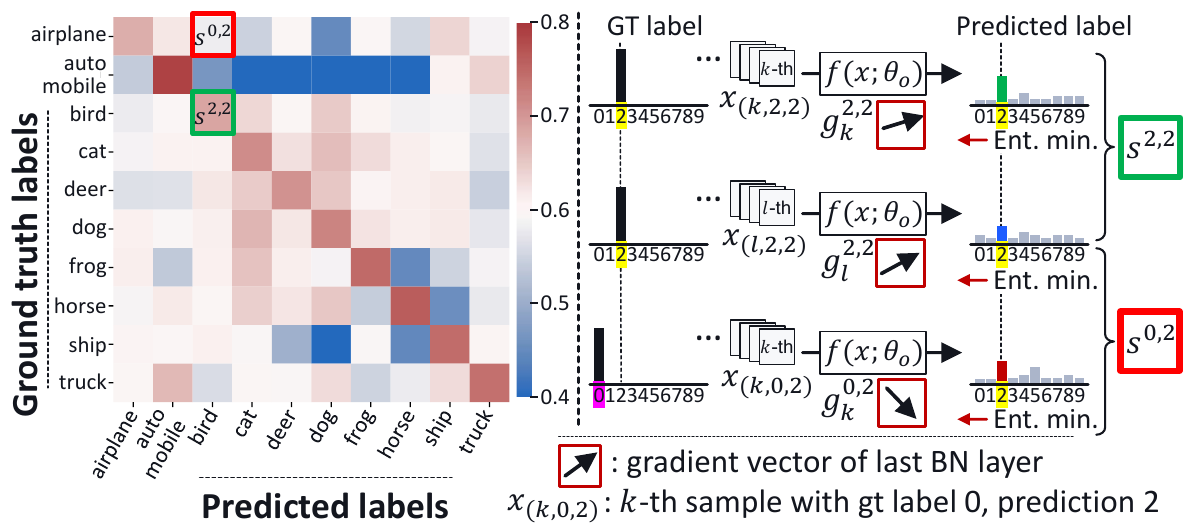}
    \vspace{-0.6cm}
    \captionof{figure}{
    Cosine similarity of gradients between samples with the same predicted label.
    We observe that wrong signals (\ie off-diagonal elements) misalign with the correct signals (\ie diagonal elements) that dominate the wisdom of crowds.}
    % While entropy minimization enforces the model to increase the confidence level of a predicted label, the cosine similarity of gradients show that correct samples share a common signal (\ie wisdom of crowds) compared to the ones from wrong samples even with the same predicted label.}
    \vspace{-0.1cm}
    \label{fig:analysis_cosine_similarity}
\end{figure}

%%%%%%%%%%% Revised version %%%%%%%%%%%
% \noindent \textbf{Decreased confidence values}
\vspace{-0.1cm}
\paragraph{Decreased confidence values}
% Fig.~\ref{fig:motivation_method_npv} shows the tendency of confidence values even with entropy minimization in test-time adaptation.
While entropy minimization enforces the model to increase the confidence value of its originally predicted label, we empirically found that wrong samples mostly show decreased values (\ie $\hat{y}^{c_o} < \Tilde{y}^{c_o}$). 
For the experiment, we perform test-time adaptation using TENT~\cite{tent} for 50 rounds using CIFAR-10-C to simulate a long-term adaptation. 
One round includes continuously changing 15 corruption types, so we repeat it 50 times without resetting the model.
With $t_{i}$ indicating the $i^{th}$ round, Fig.~\ref{fig:motivation_method_npv} (purple graph) shows that the number of samples achieving $\hat{y}^{c_o} < \Tilde{y}^{c_o}$, showing decreased confidence values, among $N$ number of test samples increases as adaptation proceeds even with the entropy minimization that enforces the model to increase its confidence value on the originally predicted label. 
In fact, the green graph in Fig.~\ref{fig:motivation_method_npv} shows that the ratio of wrong samples among the samples with decreased confidence values also increases as adaptation proceeds. 
The main reason for such an observation is due to the `wisdom of crowds', the signals learned from numerous other samples influencing the confidence level of individual samples.
% Although entropy minimization enforces the model to increase the confidence value on the predicted label of an individual sample, the individual confidence values may rise or fall, influenced by the signals that originated from numerous other predictions (\ie wisdom of crowds).
% To be more specific, although the individual signal from each sample compels the model to increase the confidence value of its own predicted label, it may be ignored by the model if the majority of other signals that are dominant show different patterns.
Specifically, although the individual signal from each sample compels the model to increase the confidence value of its own predicted label, this effect may be canceled out if other dominant signals show different patterns.
% \noindent \textbf{Wisdom of crowds from correct samples}
\vspace{-0.4cm}
\paragraph{Wisdom of crowds from correct samples}
% \edit{By utilizing a pretrained model that has a reasonable level of knowledge to capture correct signals during the adaptation, we empirically found that models generally learn the wisdom of crowds from the correct samples.}
We empirically found that models generally learn the wisdom of crowds from the correct samples.
Fig.~\ref{fig:motivation_method_histogram} demonstrates such a point with the histogram of 1) confidence values and 2) confidence difference, $\hat{y}^{c_o} - \Tilde{y}^{c_o}$, using TENT~\cite{tent} adapted for 50 rounds.
We observe that a substantial number of the samples achieving $\hat{y}^{c_o} - \Tilde{y}^{c_o} \geq 0$ are correct samples (blue). 
To be more specific, utilizing the confidence difference for distinguishing correct samples from wrong samples (red) achieves an AUROC of 89.42, which outperforms utilizing the confidence value of the adaptation model, achieving an AUROC of 58.29. 

Such an observation discloses two findings.
First, since samples achieving $\hat{y}^{c_o} \geq \Tilde{y}^{c_o}$ are mostly correct ones, the dominant signals necessary for increasing the confidence values (\ie wisdom of crowds) are originated from the correct samples.
Second, $\hat{y}^{c_o} - \Tilde{y}^{c_o}$ is an adequate metric to distinguish between correct and wrong samples. 
% We report such results with other datasets, including CIFAR-100-C~\cite{cifar} and TinyImagenet-C~\cite{imagenetC} in our Supplementary.

% propose the confidence difference between $\hat{y}^{c_o}$ and $\Tilde{y}^{c_o}$ as the novel sample selection criterion for . 
% In other words, since models capture the wisdom of crowds from the correct samples, samples achieving $\hat{y}^{c_o} - \Tilde{y}^{c_o} \geq 0$ turn out to be correct while those achieving $\hat{y}^{c_o} - \Tilde{y}^{c_o} < 0$ are mostly wrong since the signals from those samples do not align with such wisdom of crowds.

% Given a fixed column, all entropy minimization losses enforce the model to increase the predicted label but we found that wrong samples fail to increase the confidence values in Fig.~\ref{fig:motivation_method_precision_recall}.
% This is mainly due to the fact that signals from correct samples dominates the supervision to increase the confidence level of a given predicted label, so those from wrong signals are nullified by such `wisdom of crowds'. 

% \subsection{Cosine Similarity of Gradients}
% \subsection{Misaligned Wrong Signals}
% The main reason behind such an observation is due to \textit{`wisdom of crowds'} learned during training.

% \noindent \textbf{Misaligned wrong signals}
\vspace{-0.4cm}
\paragraph{Misaligned wrong signals}
We further empirically analyze why signals from wrong samples fail to increase the confidence values of the original model. 
The main reason is that signals originated from wrong samples misalign with the `wisdom of crowds' obtained from the correct samples.
For the analysis, we compute the cosine similarity of gradients between two samples with the same predicted label in Fig.~\ref{fig:analysis_cosine_similarity}.
% We use a pretrained WideResNet40-2 and obtain the gradients generated from the entropy minimization loss using CIFAR-10-C dataset. 
% We used the gradients of the weight parameters included in the last batch normalization layer in the last block since the early layers do not include class-discriminative features~\cite{swr, lim2023ttn}.
% Further details on the experiment are discussed in our Supplementary. 
For a given predicted label $i$ (column), we compute $s^{j,i}$, the cosine similarity of gradients obtained between samples of ground truth label $j$ (row) and those of predicted label $i$ as, 
% For $C$ number of target classes, we obtain the cosine similarity of ground truth label $j \in C$ and predicted label $i \in C $, 
\vspace{-0.1cm}
\begin{equation}
     s^{j,i} = \frac{1}{M_{1}M_{2}}\sum^{M_{1}}_{k=1}\sum^{M_{2}}_{l=1}  \frac{g^{j,i}_{k} \cdot g^{i,i}_{l}}{\|g^{j,i}_{k}\|\|g^{i,i}_{l}\|}, l \neq k \hspace{0.1cm} \text{if} \hspace{0.1cm} j=i,   
\end{equation}
where $g^{j,i}_{k}$ indicates the gradient vector of $k^{th}$ sample among $M_1$ number of samples with the ground truth label $j$ and the predicted label $i$, $g^{i,i}_{l}$ indicates the gradient vector of $l^{th}$ sample among $M_2$ number of samples with the ground truth label $i$ and the predicted label $i$ (\ie correct samples), and $i, j \in C$.
In other words, given a certain predicted label, we compare the gradients of the correct samples and those of the samples with the same predicted label either correct or wrong.
Thus, the diagonal elements are the results obtained by comparing the gradients between correct samples and the off-diagonal elements are obtained by comparing the gradients between correct samples and the wrong samples with the same predicted label.
We add the description of how the cosine similarity of each pair is computed on the right side of Fig.~\ref{fig:analysis_cosine_similarity}.
% where $g^{k}_{i,i}$ indicates the gradient vector of the $k^{th}$ correct sample of $M_{1}$ number of samples with ground truth label $i$, and $g^{l}_{i,j}$ indicates those of the $l^{th}$ sample predicted as $j$ among $M_{1}$ number of samples, respectively. 
% % $M_{1}$ and $M_{2}$ indicate the number of samples for each case.

Given a certain column in Fig.~\ref{fig:analysis_cosine_similarity}, all entropy minimization losses enforce the model to increase the probability value of the same predicted label.
However, we found that the signals (\ie gradients) may differ depending on the actual ground truth labels.
Specifically, the correct samples show high cosine similarity of gradients (diagonal elements, \eg $s_{2,2}$) compared to the ones with wrong samples (off-diagonal elements, \eg $s_{0,2}$). 
Since Fig.~\ref{fig:motivation_method_histogram} shows that the correct signals dominate the wisdom of crowds required for increasing the confidence value of the originally predicted label, signals that are different from these dominant signals can be suppressed and do not raise confidence values.
% signals that differ from these dominant signals lack the information necessary to increase confidence values.

% We want to clarify that the wisdom of crowds does not guarantee a model to learn with correct signals only.
We want to clarify that the wisdom of crowds does not guarantee a model to utilize the correct signals only.
Even with the wisdom of crowds, the model supervises itself with wrong predictions if the noisy losses are not filtered out. 
Such self-training with wrong knowledge significantly deteriorates the TTA performance of models, especially during the long-term adaptation~\cite{eval_cotta}.
In fact, such an issue has been widely studied in fields beyond TTA, known as the \textit{confirmation bias}~\cite{erroraccum,confirmbias,confirmbias2,dividemix,confirm_bias_noisy_label_neighbor}.
To address such an issue in TTA, we propose a sample selection method to filter out noisy samples by using the wisdom of crowds.

\begin{table*}[h]
\centering
%#################################################
% Rounds 50
%#################################################
\begin{center}
{\resizebox{1.0\textwidth}{!}{
{
\begin{tabular}{c|cc|cc|cc|cc} 
\toprule
\multirow{2}{*}{Method} & \multicolumn{2}{c|}{CIFAR-10-C}  & \multicolumn{2}{c|}{CIFAR-100-C} & \multicolumn{2}{c|}{TinyImageNet-C} & \multicolumn{2}{c}{Average}   \\
       & Closed & Open & Closed & Open & Closed & Open & Closed & Open  \\ 
\toprule
Source~\cite{wrn} & 18.27 & 18.27 & 46.75 & 46.75 & 76.71 & 76.71 & 47.24 & 47.24 \\
BN Adapt~\cite{tbn} & 14.49 & 15.73 & 39.26 & 42.67 & 61.90 & 63.00 & 38.55 & 40.47 \\
% CoTTA~\cite{cotta} & 13.76 & 82.70 & 39.45 & 97.34 & 39.45 & 97.34 \\
GCE~\cite{gce} & 43.76 & 87.94 & 44.45 & 88.69 & 97.25 & 99.00 & 61.82 & 91.88 \\
Conjugate~\cite{conjugate} & 49.57 & 92.25 & 98.97 & 98.79 & 99.38 & 99.46 & 82.64 & 96.83 \\
\hline
ENT & 87.06 & 89.26 & 56.35 & 98.76 & 99.43 & 99.50 & 80.95 & 95.84 \\
\rowcolor{Gray}+ Ours & \textbf{17.33 (-69.73)}	& \textbf{23.98 (-65.28)} & \textbf{37.69 (-18.66)}	& \textbf{40.48 (-58.28)} & \textbf{58.93 (-40.50)} & \textbf{64.01 (-35.49)} & \textbf{37.98 (-42.97)} & \textbf{42.82 (-53.02)} \\
\hline
TENT~\cite{tent} & 45.84 & 85.22 & 42.34 & 85.22 & 98.10 & 99.16 & 62.09 & 89.87 \\
\rowcolor{Gray}+ Ours & \textbf{14.10 (-31.74)} & \textbf{15.77 (-69.45)} & \textbf{38.62 (-3.72)} & \textbf{42.57 (-42.65)} & \textbf{60.87 (-37.23)} & \textbf{63.13 (-36.03)} & \textbf{37.86 (-24.23)} & \textbf{40.49 (-49.38)} \\
\hline
EATA~\cite{eata} & 29.78 & 82.05 & 49.31 & 98.75 & 59.82 & 63.47 & 46.30 & 81.42 \\
\rowcolor{Gray}+ Ours & \textbf{14.07 (-15.71)} & \textbf{15.65 (-66.40)} & \textbf{38.44 (-10.87)} & \textbf{42.47 (-56.28)} & \textbf{59.80 (-0.02)} & \textbf{62.08 (-1.39)} & \textbf{37.44 (-8.86)} & \textbf{40.07 (-41.35)} \\
\hline
SWR~\cite{swr} & 10.21 & 90.55 & 35.78 & 73.05 & 62.39 & 76.13 & 36.13 & 79.91 \\
\rowcolor{Gray}+ Ours & \textbf{10.12 (-0.09)} & \textbf{72.58 (-17.97)} & \textbf{35.64 (-0.14)} & \textbf{45.68 (-27.37)} & \textbf{55.15 (-7.24)} & \textbf{61.91 (-14.22)} & \textbf{33.64 (-2.49)} & \textbf{60.06 (-19.85)} \\
\toprule
\end{tabular}}}}
\end{center}
\vspace{-0.6cm}
\caption{Error rates of image classification after 50 rounds of adaptation (\ie long-term test-time adaptation). We note the performance gain by reduced error rates.}
\label{tab:main_image_classifcation_rounds50} 
\vspace{-0.2cm}
\end{table*}
%##################################################################################################

%#################################################################################################
\begin{table*}[h]
\centering
%#################################################
% Rounds 100
%#################################################
\begin{center}
{\resizebox{1.0\textwidth}{!}{
{
\begin{tabular}{c|cc|cc|cc|cc} 
\toprule
\multirow{2}{*}{Method} & \multicolumn{2}{c|}{CIFAR-10-C}  & \multicolumn{2}{c|}{CIFAR-100-C} & \multicolumn{2}{c|}{TinyImageNet-C} & \multicolumn{2}{c}{Average}   \\
       & Closed & Open & Closed & Open & Closed & Open & Closed & Open  \\ 
\toprule
Source~\cite{wrn} & 18.27 & 18.27 & 46.75 & 46.75 & 76.71 & 76.71 & 47.24 & 47.24 \\
BN Adapt~\cite{tbn} & 14.49 & 15.73 & 39.26 & 42.67 & 61.90 & 63.00 & 38.55 & 40.47 \\
% CoTTA~\cite{cotta} & 13.76 & 82.70 & 39.45 & 97.34 & 39.45 & 97.34 \\
GCE~\cite{gce} & 12.81 & 25.70 & 35.83 & 45.78 & 62.84 & 71.41 & 37.16 & 47.63 \\
Conjugate~\cite{conjugate} & 12.84 & 24.96 & 36.67 & 81.19 & 82.83 & 92.66 & 44.11 & 66.27 \\
\hline
ENT & 16.30 & 47.54 & 38.74 & 58.16 & 79.69 & 91.74 & 44.91 & 65.81 \\
\rowcolor{Gray}+ Ours & \textbf{13.41 (-2.89)} & \textbf{16.93 (-30.61)} & \textbf{37.55 (-1.19)} & \textbf{42.60 (-15.56)} & \textbf{63.89 (-15.80)} & \textbf{69.01 (-22.73)} & \textbf{38.28 (-6.63)} & \textbf{42.85 (-22.96)} \\
\hline
TENT~\cite{tent} & 12.56 & 27.80 & \textbf{36.04} & 45.26 & 68.53 & 80.93 & 39.04 & 51.33 \\
\rowcolor{Gray}+ Ours & \textbf{12.39 (-0.17)} & \textbf{14.94 (-12.86)} & 36.18 (+0.14) & \textbf{39.62 (-5.64)} & \textbf{59.90 (-8.63)} & \textbf{63.31 (-17.62)} & \textbf{36.16 (-2.88)} & \textbf{39.29 (-12.04)} \\
\hline
EATA~\cite{eata} & 12.39 & 25.52 & 36.39 & 54.22 & \textbf{59.02} & \textbf{61.72} & \textbf{35.93} & 47.15 \\
\rowcolor{Gray}+ Ours & \textbf{12.35 (-0.04)} & \textbf{14.92 (-10.60)} & \textbf{36.25 (-0.14)} & \textbf{39.58 (-14.64)} & 59.30 (+0.28) & 62.11 (+0.39) & 35.97 (+0.04) & \textbf{38.87 (-8.28)} \\
\hline
SWR~\cite{swr} & 10.76 & 29.32 & \textbf{34.21} & 44.79 & 60.34 & 65.18 & 35.10 & 46.43 \\
\rowcolor{Gray}+ Ours & \textbf{10.74 (-0.02)} & \textbf{27.52 (-1.80)}  & 34.23 (+0.02) & \textbf{41.52 (-3.27)} & \textbf{58.50 (-1.84)} & \textbf{62.94 (-2.24)} & \textbf{34.49 (-0.61)} & \textbf{44.00 (-2.43)} \\
\toprule
\end{tabular}}}}
\end{center}
\vspace{-0.58cm}
\caption{Error rates of image classification after 1 round of adaptation (\ie short-term test-time adaptation). We note the performance gain by reduced error rates.}
\vspace{-0.42cm}
\label{tab:main_image_classifcation_rounds1}
\end{table*}
%##################################################################################################

\subsection{Proposed Method}
\label{sec3:sample_selection}

% As aforementioned, our empirical finding is that models achieve $f(\theta_a; \hat{y_o})$ lower than $f(\theta_a; \hat{y_o})$ for the noisy samples, samples with wrong predictions or those from unknown classes. 
% On the other hand, models achieve $f(\theta_a; \hat{y_o})$ higher than $f(\theta_a; \hat{y_o})$ for the correct samples. 
% The main reason is that such noisy samples do not share a common pattern to increase the probability value on a certain target.
% For example, given the predicted label `dog', gradients generated from correct samples share a common pattern to increase the probability value on target class `dog'.
% However, misclassified samples which are predicted as dogs (e.g., cats or horses) do not share a common pattern to increase the probability value on `dog' since their input images are different among each other. 
As shown in Fig.~\ref{fig:method}, we propose a \textit{simple yet effective} sample selection method using the confidence difference between $\Tilde{y}^{c_o}$ and $\hat{y}^{c_o}$.
Our sample selection criterion is formulated as
% \begin{equation}
%      \Phi(x; \theta_a, \theta_o) = \mathbb{1}_{\hat{y}^{c_o} \geq \Tilde{y}^{c_o}}(x),
% \end{equation}
\begin{equation}
     \Phi(\hat{y}^{c_o}, \Tilde{y}^{c_o}) = \mathbb{1}\left(\hat{y}^{c_o} \geq \Tilde{y}^{c_o}\right),
\end{equation}
where $\Phi(\cdot)$ is our sample selection criterion and $\mathbb{1}(\cdot)$ is the indicator function. 
% By using our selection criterion, the total objective function for using entropy minimization is formulated as 
Our total objective function using entropy minimization is formulated as 
\begin{equation}
    \mathcal{L}^{\text{main}}(x; \theta_{a}) = \Phi(\hat{y}^{c_o}, \Tilde{y}^{c_o}) \cdot H(\hat{y_i}) - \lambda_{\text{max}} H(\overline{y}).
\end{equation}
$H(p)=\Sigma^{C}_{k=1}p^{k}\log{p^{k}}$, $\overline{y} = \frac{1}{N}\Sigma^{C}_{k=1}\hat{y_i}$, and $\lambda_{max}$ is the scalar value for balancing the two loss values. 
Note that $H(\overline{y})$ has been widely used in previous studies~\cite{swr, lim2023ttn, discrimnative_clustering, shot, ssl_support, adacontrast} to prevent the model from making imbalanced predictions towards a certain class.

Recent studies require the pre-deployment stage that obtains the necessary information needed for each method by using the samples from the source data before the adaptation phase~\cite{swr, eata, lim2023ttn}. 
However, we want to emphasize that our method does not require such a pre-deployment stage as well as those samples from the source distribution.
Due to such an advantage, our method can be easily applied to existing TTA methods without additional preparations.
Through extensive experiments, we demonstrate the wide applicability of our method to existing TTA methods. 
% Since TTA studies~\cite{tent, eata, swr, conjugate, adacontrast, cafa} widely utilize entropy minimization loss, our method can be applied to existing TTA methods and improve their performances.  

\vspace{-0.1cm}
\section{Experiments}
\begin{table*}[h!]
\centering
%#################################################
% Rounds 100
%#################################################
\vspace{-0.1cm}
\subfloat[Average mIoU after the adaptation of each domain.\label{tab:continual_segmentation_quant}]
{
\centering
\begin{minipage}{0.63\linewidth}{
\begin{center}
{\resizebox{\textwidth}{!}{
{

% Method & BDD-100K & Mapillary & GTAV & SYNTHIA & Averag

\begin{tabular}{l|ccc|c} \toprule
Time & \multicolumn{3}{l|}{$t\xrightarrow{\hspace*{7cm}}$} & \multirow{2}{*}{Average} \\ 
\cmidrule(r){0-3}
Method & \multicolumn{1}{c}{Cityscapes} & \multicolumn{1}{c}{BDD-100K} & \multicolumn{1}{c|}{Mapillary} &  \\ \toprule
Source~\cite{deeplab}                & 34.74 & 16.15 & 36.97 & 29.29 \\
BN Adapt~\cite{tbn}   & 40.77 & 25.21 & 39.10 & 35.03 \\
TTN~\cite{lim2023ttn}        & 46.28 & 28.07 & 45.46 & 39.94\\
\hline
TENT~\cite{tent}      & 46.73 & 29.59 & 35.69 & 37.34 \\
\rowcolor{Gray}+ Ours & \textbf{46.76 (+0.03)} & \textbf{30.55 (+0.96)} & \textbf{43.42 (+7.73)} & \textbf{40.24 (+2.90)} \\
\hline
SWR~\cite{swr} & 46.17 & 10.70 & 1.28 & 19.38 \\
\rowcolor{Gray}+ Ours & \textbf{46.65 (+0.48)} & \textbf{32.28 (+21.58)} & \textbf{45.09 (+43.81)} & \textbf{41.34 (+21.96)} \\ \hline
\end{tabular}
}}}
\end{center}
\vspace{-1.em}
}
\end{minipage}
}
%#################################################
% Round 1
%#################################################
% \subfloat[mIoU changes as adaptation steps.\label{tab:continual_segmentation_quali}]{
\subfloat[mIoU changes during adaptation\edit{$^2$}.\label{tab:continual_segmentation_quali}]{
\centering
\begin{minipage}{0.33\linewidth}{
\begin{center}
\vspace{-0.1cm}
\includegraphics[width=0.92\linewidth]{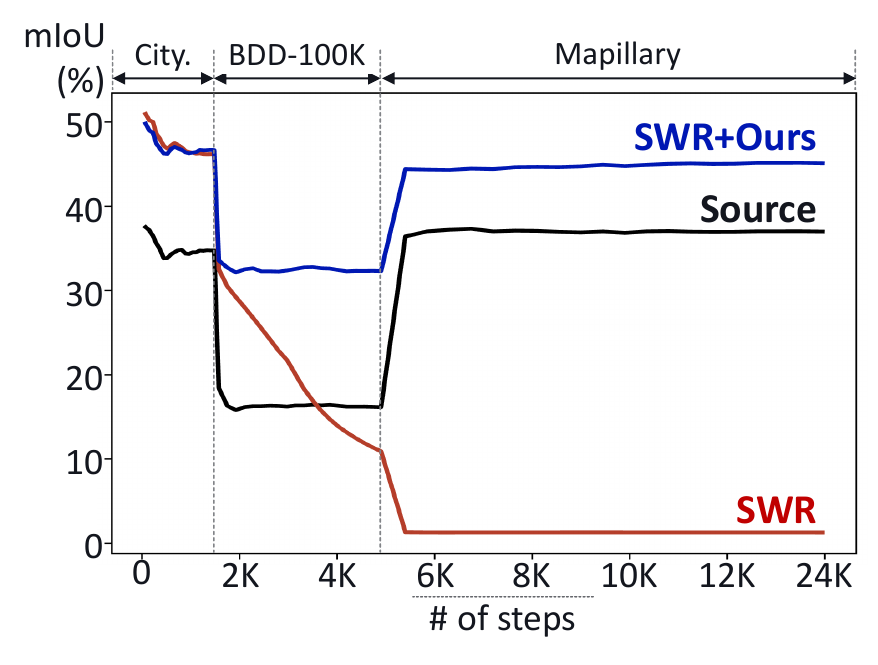}
\vspace{-0.35cm}
\end{center}
}\end{minipage}
}
\vspace{-0.25cm}
\caption{Semantic segmentation performance (mIoU) on continuously-changing target domains with 1 round of adaptation. We evaluate with DeepLabV3Plus-ResNet-50~\cite{deeplab} pretrained on GTAV dataset.}
% Our method brings larger performance gains as adaptation proceeds.}
% We use the train sets of target domains to simulate a long-term adaptation.

\label{tab:main_semantic_segmentation_lifelong} 
\vspace{-0.15cm}
\end{table*}
%##################################################################################################

%#################################################################################################
\begin{table*}[h]
\centering
%#################################################
% Rounds 100
%#################################################
\begin{center}
{\resizebox{1.0\textwidth}{!}{
{

% Method & BDD-100K & Mapillary & GTAV & SYNTHIA & Averag

\begin{tabular}{c|cc|cc|cc|cc|cc} 
\toprule
\multirow{2}{*}{Method} & \multicolumn{2}{c|}{BDD-100K}  & \multicolumn{2}{c|}{Mapillary} & \multicolumn{2}{c|}{GTAV} & \multicolumn{2}{c|}{SYNTHIA} & \multicolumn{2}{c}{Average}   \\
& Round 1 & Round 10 & Round 1 & Round 10 & Round 1 & Round 10 & Round 1 & Round 10 & Round 1 & Round 10  \\ 
\toprule
Source~\cite{deeplab} & 43.50 & 43.50 & 54.37 & 54.37 & 44.55 & 44.55 & 22.78 & 22.78 & 41.30 & 41.30 \\
BN Adapt~\cite{tbn} & 43.60 & 43.60 & 47.66 & 47.66 & 43.22 & 43.22 & 25.72 & 25.72 & 40.05 & 40.05 \\
TTN~\cite{lim2023ttn} & 48.43 & 48.43 & 57.28 & 57.28 & 46.71 & 46.71 & 26.41 & 26.41 & 44.71 & 44.71 \\
\hline
TENT~\cite{tent} & 48.90 & 47.57 & 57.94 & 53.36 & 48.14 & 17.91 & 26.88 & 13.36 & 45.47 & 33.05 \\
\rowcolor{Gray}+ Ours & 48.90 & \textbf{48.88 (+1.31)} & 57.94 & \textbf{56.49 (+3.13)} & \textbf{48.28 (+0.14)} & \textbf{47.98 (+30.07)} & \textbf{26.90 (+0.02)} & \textbf{25.62 (+12.26)} & \textbf{45.51 (+0.04)} & \textbf{44.74 (+11.69)} \\
\hline
SWR~\cite{swr} & 49.39 & 49.68 & \textbf{59.33} & \textbf{59.70} & 47.82 & 48.13 & \textbf{28.40} & 1.18 & 46.24 & 39.67 \\
\rowcolor{Gray}+ Ours & \textbf{49.88 (+0.49)} & \textbf{50.57 (+0.89)} & 58.79 (-0.54) & 58.89 (-0.81) & \textbf{49.17 (+1.35)} & \textbf{49.27 (+1.14)} & 27.75 (-0.65) & \textbf{27.82 (+26.64)} & \textbf{46.40 (+0.16)} & \textbf{46.64 (+6.97)} \\
\toprule
\end{tabular}}}}
\end{center}
\vspace{-0.5cm}
\caption{Semantic segmentation performance (mIoU) on a fixed target domain with 10 rounds of adaptation. We use DeepLabV3Plus-ResNet-50~\cite{deeplab} pretrained on Cityscapes dataset.}
\vspace{-0.5cm}
\label{tab:main_semantic_segmentation_single} 
\end{table*}

\subsection{Experimental Setup}
\noindent \textbf{Datasets}
For the image classification task, we use the widely used corruption benchmark datasets: CIFAR-10/100-C and TinyImageNet-C.
We apply 15 different types of corruptions (e.g., gaussian noise) to CIFAR-10/100~\cite{cifar} and TinyImageNet~\cite{tinyimagenet}.
Pretrained models are trained on the clean train set and adapted to the corrupted test set. 
For the open-set setting, we use SVHN~\cite{svhn} for CIFAR-10/100-C, and ImageNet-O~\cite{imagenet_o} for TinyImagenet-C, where we apply the same corruption type as the original test sets. 
We term the datasets as SVHN-C and ImageNet-O-C, respectively. 
We apply the identical corruption type in order to construct open-set samples that are drawn from the same domain shift but with unknown classes.
For the semantic segmentation task under continually changing domains, we use a model pretrained on GTAV~\cite{gta5}, and evaluate it with 
Cityscapes~\cite{cityscapes}, BDD-100K~\cite{bdd100k}, and Mapillary~\cite{mapillary}. 
For semantic segmentation with a fixed target domain with multiple rounds, we use the Cityscapes for the source distribution and BDD-100K~\cite{bdd100k}, GTAV~\cite{gta5}, Mapillary~\cite{mapillary}, and SYNTHIA~\cite{synthia} for the target distributions. 
Note that the semantic segmentation task inherently includes open-set classes in the test set (e.g., traffic cones in BDD100K not shown during training with Cityscapes).

% \noindent \textbf{Evaluation Settings}
\vspace{-0.6cm}
\paragraph{Evaluation settings}
Following the recent TTA studies, we evaluate TTA models under continuously changing domains without resetting the model after each domain~\cite{cotta, lim2023ttn, eata}.
For the closed-set and open-set continual long-term TTA in the image classification, we perform adaptation for 50 rounds to simulate a long-term TTA with continuously changing domains. 
We report both TTA performances after 1 round (\ie short-term TTA) and 50 rounds (\ie long-term TTA). 
Note that we evaluate predictions made during online model adaptation, not after visiting the
entire test set, strictly following the established TTA settings. 
For the open-set TTA, we construct the mini-batch that includes an equal number of closed-set samples (e.g., CIFAR-10-C, shot noise) and open-set samples (e.g., SVHN-C, shot noise). 
Although included in the mini-batch, we exclude open-set samples from the evaluation and only evaluate models with closed-set samples. 
To the best of our knowledge, our work is the first paper to conduct experiments with the open-set TTA. 
We report the error rates and mean intersection of union (mIoU) for image classification and semantic segmentation, respectively.  

% \noindent \textbf{Baselines}
\vspace{-0.5cm}
\paragraph{Baselines}
We mainly compare our method with previous methods addressing noisy labels~\cite{gce} or improving pseudo-labeling performances in TTA~\cite{conjugate,eata}.
Note that ENT denotes updating all parameters while TENT~\cite{tent} only updates affine parameters of the batch normalization layers, both utilizing the entropy minimization loss function.
Gray-shaded digits indicate the performance gain by applying our method to each baseline model, and bold digits indicate the better performance between the two methods.

% \noindent \textbf{Implementation Details}
\vspace{-0.5cm}
\paragraph{Implementation details}
For the image classification, we use the learning rate of $1e$-$3$ and $1e$-$4$ for models updating only affine parameters (TENT~\cite{tent}, EATA~\cite{eata}, GCE~\cite{gce}, Conjugate~\cite{conjugate}) and all parameters (ENT, SWR~\cite{swr}), respectively.
We use the batch size of 200 and Adam optimizer~\cite{adam} for all experiments. 
For experiments conducting small batch sizes in Table~\ref{tab:main_hyper_param}, we use the learning rate of $1e$-$4$ and update models after 200 steps, following TTN~\cite{lim2023ttn}.
For the semantic segmentation, we use the learning rate of $1e$-$6$ and batch size of 2 following TTN.
Regarding using TTN in semantic segmentation, we update the test batch statistics in an online manner to further improve the segmentation performance for all experiments. 
Further details on our experimental setup are included in our supplementary.

% \begin{table*}[h]
% \centering
% \begin{center}
% {\resizebox{1.0\textwidth}{!}{
% {

% \begin{tabular}{c|c|cc|cc} 
% \toprule
% Adaptation Method & OoD Method & \multicolumn{2}{c|}{CIFAR10 / SVHN-C} & \multicolumn{2}{c|}{CIFAR100 / SVHN-C}  \\
% \multirow{4}{*}{Tent~\cite{tent}} & & AUROC & FPR & AUROC & FPR   \\
% & MSP & - & - & - & -  \\ 
% & Max Logit & - & - & - & -  \\ 
% & Energy & - & - & - & -  \\ 
% & Ours & - & - & - & -  \\ 
% \toprule
% \end{tabular}}}}
% \end{center}
% \vspace{-.5em}
% \label{tab:main_open_set} \vspace{-1.em}
% \caption{OoD Evaluation}
% \end{table*}

%#################################################################################################
\begin{table*}[h!]
\centering
%#################################################
% Rounds 100
%#################################################
\vspace{-0.1cm}
\subfloat[
Negative samples including closed-set wrong samples.
\label{tab:ood_include}
]{
\centering
\begin{minipage}{0.48\linewidth}{
\begin{center}
{\resizebox{1.0\textwidth}{!}{
{

\begin{tabular}{c|cc|cc} 
\toprule
\multirow{2}{*}{Method} & \multicolumn{2}{c|}{CIFAR-10 / SVHN-C} & \multicolumn{2}{c}{CIFAR-100 / SVHN-C}  \\
& AUROC$\uparrow$ & FPR@TPR95$\downarrow$ & AUROC$\uparrow$ & FPR@TPR95$\downarrow$ \\
\hline
MSP~\cite{msp_baseline} & 51.87 & 92.39 & 60.69 & 87.96  \\ 
Max Logit~\cite{maxlogit} & 54.68 & 90.31 & 64.88 & 85.45  \\ 
Energy~\cite{energy} & 54.68 & 90.30 & 64.87 & 85.46  \\ 
 Ours & \textbf{88.24} & \textbf{40.34} & \textbf{83.76} & \textbf{64.86}  \\ 
\toprule
\end{tabular}}}}
\end{center}
\vspace{-1.em}
}
\end{minipage}
}
%#################################################
% Round 1
%#################################################
\vspace{-0.1cm}
\subfloat[
Negative samples excluding closed-set wrong samples.
\label{tab:ood_exclude}
]{
\centering
\begin{minipage}{0.48\linewidth}{
\begin{center}
{\resizebox{1.0\textwidth}{!}{
{

\begin{tabular}{c|cc|cc} 
\toprule
\multirow{2}{*}{Method} & \multicolumn{2}{c|}{CIFAR-10 / SVHN-C} & \multicolumn{2}{c}{CIFAR-100 / SVHN-C}  \\
& AUROC$\uparrow$ & FPR@TPR95$\downarrow$ & AUROC$\uparrow$ & FPR@TPR95$\downarrow$ \\
\hline
MSP~\cite{msp_baseline} & 50.83 & 93.64 & 56.14 & 90.34  \\ 
Max Logit~\cite{maxlogit} & 56.25 & 90.65 & 62.76 & 87.35  \\ 
Energy~\cite{energy} & 56.26 & 90.63 & 62.79 & 87.27  \\ 
Ours & \textbf{83.50} & \textbf{54.46} & \textbf{82.17} & \textbf{73.16}  \\ 
\toprule
\end{tabular}}}}
\end{center}
\vspace{-1.em}
}
\end{minipage}
}
\vspace{-0.1cm}
\caption{Utilizing the confidence difference for thresholding in open-set test time adaptation. We use TENT~\cite{tent} adapted to each target domain including open-set classes (SVHN-C) for 50 rounds.}
\label{tab:ood_evaluation} 
\vspace{-0.35cm}
\end{table*}
%##################################################################################################
\begin{table}[t]
%#################################################
% Rounds 100
%#################################################
\vspace{-0.05cm}
{\centering

\begin{center}
{\resizebox{0.45\textwidth}{!}{
{
\begin{tabular}{c|cc|cc} 
\toprule
\multirow{2}{*}{Method} & CIFAR-10-C & CIFAR-100-C & \multicolumn{2}{c}{CIFAR-10/100-C} \\
& Error Rate (\%) & Error Rate (\%) & Memory (MB) & Time (ms) \\
\hline
ENT                         & 88.16 & 77.56 & 1147 & 22.98  \\ 
SWR~\cite{swr}              & 50.38	& 54.42 & 1155 & 47.97  \\ 
TENT~\cite{tent}            & 65.53 & 63.78 & 556 & 18.38  \\ 
EATA~\cite{eata}            & 55.92 & 74.03 & 559 & 37.04 \\ 
\rowcolor{Gray} TENT~\cite{tent} + Ours       & \textbf{14.94} & \textbf{40.60} & 565 & 26.62  \\ 

\hline
\toprule
\end{tabular}}}}
\end{center}
\vspace{-0.5cm}
}
\caption{Comparisons on error rates (\%), memory (MB), and time (ms). 
% For the error rates, we average the long-term adaptation performance of the closed-set and open-set error rates for each dataset. 
For the time, we report the average time after 5000 trials on NVIDIA RTX A5000.}
\vspace{-0.6cm}
\label{tab:main_inference} 
\end{table}

% \begin{table}[t]
% %#################################################
% % Rounds 100
% %#################################################
% \vspace{-0.2cm}
% {\centering

% \begin{center}
% {\resizebox{0.49\textwidth}{!}{
% {
% \begin{tabular}{c|ccc|cc} 
% \toprule
% \multirow{2}{*}{Method} & CIFAR-10-C & CIFAR-100-C & CIFAR-10/100-C & \multicolumn{2}{c}{Tiny-Imagenet-C}  \\
% & Error Rate (\%) & Error Rate (\%) & Time (ms) & Error Rate (\%) & Time (ms) \\
% \hline
% ENT                         & 88.16 & 77.56	& 22.98 & 99.47 & 34.15  \\ 
% SWR \& NSP~\cite{swr}       & 50.28	& 54.42 & 47.97 & 69.26 & 79.13  \\ 
% EATA~\cite{eata}          & 55.92 & 74.03 & 37.04 & 61.65 & 46.32 \\ 
% TENT~\cite{tent}            & 65.53 & 63.78 & 18.38 & 98.63 & 31.78  \\ 
% + Ours                      & 14.94 & 40.60 & 26.62 & 62.00 & 37.32  \\ 

% \hline
% \toprule
% \end{tabular}}}}
% \end{center}
% \vspace{-0.5cm}
% }
% \caption{Comparisons on test-time adaptation time (ms). We use ResNet50~\cite{resnet} and report the time of inference and backpropagation. We report the average time after 5000 trials on NVIDIA RTX A5000.}
% \vspace{-0.65cm}
% \label{tab:main_inference} 
% \end{table}

%#################################################################################################
\begin{table*}[t]
\centering
%#################################################
% Rounds 100
%#################################################
\subfloat[
Long-term adaptation
]{
\centering
\begin{minipage}{0.5\linewidth}{
\begin{center}
{\resizebox{1.0\textwidth}{!}{
{

\begin{tabular}{c|cc|cc} 
\toprule
\multirow{2}{*}{Method} & \multicolumn{2}{c|}{ResNet50~\cite{resnet}}  & \multicolumn{2}{c}{WDR28~\cite{wrn}} \\
& Closed & Open & Closed & Open \\ 
\toprule
Source                & 48.80 & 48.80 & 43.52 & 43.52 \\
BN Adapt~\cite{tbn}   & 16.01 & 16.89 & 20.43 & 23.61 \\
\hline
TENT~\cite{tent}      & 61.69 & 83.62 & 56.00 & 77.72 \\
\rowcolor{Gray}+ Ours & \textbf{15.28 (-46.41)} & \textbf{16.99 (-66.63)} & \textbf{20.16 (-35.84)} & \textbf{23.70 (-54.02)} \\
\hline
SWR~\cite{swr} & 16.19 & 88.53 & 17.94 & 90.15 \\
\rowcolor{Gray}+ Ours & \textbf{16.08 (-0.11)} & \textbf{71.83 (-16.70)} & \textbf{15.35 (-2.59)} & \textbf{83.76 (-6.39)} \\
\hline
\toprule
\end{tabular}}}}
\end{center}
\vspace{-1.5em}
}
\end{minipage}
}
%#################################################
% Round 1
%#################################################
\subfloat[
Short-term adaptation
\label{tab:cifar10}
]{
\centering
\begin{minipage}{0.475\linewidth}{
\begin{center}
{\resizebox{1.0\textwidth}{!}{
{

\begin{tabular}{c|cc|cc} 
\toprule
\multirow{2}{*}{Method} & \multicolumn{2}{c|}{ResNet50~\cite{resnet}}  & \multicolumn{2}{c}{WDR28~\cite{wrn}} \\
& Closed & Open & Closed & Open \\ 
\toprule
Source                & 48.80 & 48.80 & 43.52 & 43.52 \\
BN Adapt~\cite{tbn}   & 16.01 & 16.89 & 20.43 & 23.61 \\
\hline
TENT~\cite{tent}      & 14.03 & 22.76 & \textbf{18.23} & 32.74 \\
\rowcolor{Gray}+ Ours & \textbf{13.82 (-0.21)} & \textbf{16.36 (-6.40)} & 18.32 (+0.09) & \textbf{23.40 (-9.34)} \\
\hline
SWR~\cite{swr} & 13.81 & 45.58 & 16.62 & 83.08 \\
\rowcolor{Gray}+ Ours & \textbf{13.80 (-0.01)} & \textbf{43.35 (-2.23)} & \textbf{15.73 (-0.89)} & \textbf{75.89 (-7.19)}  \\
\hline
\toprule
\end{tabular}}}}
\end{center}
\vspace{-1.5em}
}
\end{minipage}
}
\vspace{-0.25cm}
\caption{Error rates of image classification on CIFAR-10-C using diverse architectures.} 
% We consistently improve the TTA performance of the baseline models with diverse architectures.}
\vspace{-0.23cm}
\label{tab:main_model_agnostic}
\end{table*}
%##################################################################################################
%#################################################################################################
\begin{table*}[t]
\centering
%#################################################
% Rounds 100
%#################################################
\subfloat[
Robustness to learning rates
\label{tab:main_learning_rate}
]{
\centering
\begin{minipage}{0.49\linewidth}{
\begin{center}
{\resizebox{0.8\textwidth}{!}{
{

\begin{tabular}{c|cccc|c} 
\toprule
\multirow{2}{*}{Method} & \multicolumn{4}{c|}{Learning rate}  & \multirow{2}{*}{Std.$\downarrow$} \\
& 0.005 & 0.001 & 0.0005 & 0.0001 &  \\ 
\toprule
Source & 76.71 & 76.71 & 76.71 & 76.71 & 0 \\
TENT~\cite{tent} & 99.51 & 89.91 & 75.02 & 63.83 & 15.79 \\
\rowcolor{Gray}+ Ours & \textbf{64.14} & \textbf{60.04} & \textbf{59.59} & \textbf{58.76} & \textbf{2.40} \\
\hline
\toprule
\end{tabular}}}}
\end{center}
\vspace{-.5em}
\label{tab:main_cifar} \vspace{-1.em}
}
\end{minipage}
}
%#################################################
% Round 1
%#################################################
\subfloat[
Robustness to batch sizes
\label{tab:main_batch_size}
]{
\centering
\begin{minipage}{0.49\linewidth}{
\begin{center}
{\resizebox{0.8\textwidth}{!}{
{

\begin{tabular}{c|cccc|c} 
\toprule
\multirow{2}{*}{Method} & \multicolumn{4}{c|}{Batch size}  & \multirow{2}{*}{Std.$\downarrow$} \\
& 64 & 32 & 16 & 8 &  \\ 
\toprule
Source & 76.71 & 76.71 & 76.71 & 76.71 & 0 \\
TENT~\cite{tent} & 67.54 & 72.62 & 81.21 & 94.75 & 11.90 \\
\rowcolor{Gray}+ Ours & \textbf{60.32} & \textbf{62.14} & \textbf{65.88} & \textbf{73.83} & \textbf{5.99} \\
\hline
\toprule
\end{tabular}}}}
\end{center}
\vspace{-.5em}
\vspace{-1.em}
}
\end{minipage}
}
\vspace{-0.26cm}
\caption{Error rates of image classification on TinyImageNet-C with diverse learning rates and batch sizes. Std. is the abbreviation of the standard deviation.}
% We use Tiny-ImageNet-C for the evaluation. Our method shows a stable performance with low standard deviations compared to TENT~\cite{tent}.
\label{tab:main_hyper_param} 
\vspace{-0.53cm}
\end{table*}
%##################################################################################################

\subsection{Results}
\vspace{-0.2cm}
\noindent \textbf{Image classification}
\blfootnote{\hspace*{-0.18cm}$^2$Note that the performance variation of the source model in Cityscapes is due to the order of data samples (\eg challenging ones in the later stage), not due to the adaptation.}
As shown in Table~\ref{tab:main_image_classifcation_rounds50}, existing TTA models show a large performance degradation during the long-term adaptation.
This is mainly due to the confirmation bias, caused by the unsupervised losses that inevitably include noisy losses.
We significantly improve the long-term performance of the existing four different TTA models in both closed-set and open-set TTA. 
For example, we improve the error rate of TENT~\cite{tent} by an average of 24.23\% and 49.38\% in the closed-set and open-set settings, respectively.
% Such results demonstrate that our method can effectively filter out both samples with mispredictions and those with unknown classes.
% This indicates that wrong samples and open-set samples show decreased confidence values since they fail to meet with the wisdom of crowds found in the correct samples. 
Note that we do not use prior knowledge of whether the target distribution includes open-set samples or not. 
Additionally, Table~\ref{tab:main_image_classifcation_rounds1} shows that our method also generally improves the short-term TTA performances.  

While previous studies focused on improving the performance of closed-set TTA until now, our results show that they suffer from a large performance drop when adapted with open-set classes included. 
We believe that this is a practical setting since we can not guarantee that samples from the target distributions are always drawn from the classes learned during the training stage. 
Such results indicate that improving the TTA performance with open-set classes is yet to be explored in the future. 

\noindent \textbf{Semantic segmentation}
Table~\ref{tab:main_semantic_segmentation_lifelong} shows the semantic segmentation performance with continuously changing domains.
We evaluated a model pretrained on GTAV~\cite{gta5} with real-domain datasets (Cityscapes~\cite{cityscapes}, BDD-100K~\cite{bdd100k}, and Mapillary~\cite{mapillary}) in order to simulate the situation where real-world target datasets are not available with only synthetic datasets provided.
We observe that the performance gain by applying our method increases as the adaptation proceeds.
% Similar to image classification, the main reason is due to the significant performance drop observed during long-term adaptation. 
For example, SWR~\cite{swr} (Table~\ref{tab:continual_segmentation_quali} - red) suffers from a large performance drop with the last target domain, Mapillary (1.28 mIoU), while ours (Table~\ref{tab:continual_segmentation_quali} - blue) shows a stable level of performance (45.09 mIoU). 
% For Table~\ref{tab:continual_segmentation_quali}, we visualize the average mIoU  evaluate models after certain steps and show the average mIoU up to then. 
Regarding Table~\ref{tab:continual_segmentation_quali}, we evaluate models after certain steps and show the average mIoU up to then.
While the model without adaptation (\ie source) does not suffer from the error accumulation, it fails to bring performance gain.
On the other hand, our method not only brings performance gain but also circumvents error accumulation by filtering the noisy losses.
% \footnote{Note that the performance variation of the source model in Cityscapes is due to the order of data samples (\eg challenging ones in the later stage), not due to the adaptation.}.

% Our method effectively mitigates such an issue.

Table~\ref{tab:main_semantic_segmentation_single} also reports the semantic segmentation performance with a fixed target domain over multiple rounds of adaptation. 
We observe that applying our method improves the performance of TENT~\cite{tent} and SWR~\cite{swr} by an average of 11.69 mIoU and 6.97 mIoU, respectively, after 10 rounds.
As aforementioned, performing test-time adaptation in semantic segmentation needs to address not only the wrong predictions but also the inherently included open-set classes in the target distribution. 
Our method again improves TTA performance by effectively discarding such noisy pixels.
We believe such a filtering mechanism is especially important in safety-critical applications in two aspects.
First, it prevents the performance drop caused by learning with noisy losses. 
Second, when confronted with unknown objects, we could alarm a device immediately, which could be the starting point for it to take a different action (e.g., autonomous vehicles swerving directions to avoid running into wild animals unexpectedly shown on roads) ~\cite{sml}.

\vspace{-0.2cm}
\section{Further Analysis}
\subsection{Utilizing Confidence Difference as Thresholds}
\vspace{-0.1cm}
We show that the confidence difference is an adequate metric to differentiate between correct samples and noisy samples, given that a pretrained model is adapting to a novel domain.
For the evaluation, we train TENT~\cite{tent} and compare utilizing confidence difference as the thresholding metric with existing prediction-based out-of-distribution (OoD) methods~\cite{msp_baseline,maxlogit,energy}.
By setting the correct samples as the positive samples, we analyze two different negative samples: negative samples 1) including closed-set wrong samples and 2) excluding closed-set wrong samples.
The former case shows how well a given metric differentiates between correct samples and noisy samples, including both closed-set and open-set samples.
The latter case evaluates how well a given metric distinguishes between the correct samples and open-set samples only. 
Table~\ref{tab:ood_evaluation} shows that using confidence difference outperforms the existing OoD metrics in both cases. 
In addition to the superior performance, another advantage of using the confidence difference is that we can filter the noisy samples immediately, while the existing OoD metrics need the entire test samples in order to choose the threshold with the best AUROC score. 
Such a result indicates that confidence difference can also be widely used to distinguish out-of-distribution samples in future studies with adapted models.
% Such a result indicates that confidence difference can also be widely used to distinguish out-of-distribution samples in future studies given that a pretrained model is adapting to a target domain. 

\vspace{-0.1cm}
% \subsection{Comparisons on Test-Time Adaptation Time}
% \vspace{-0.1cm}
% Along with the TTA performances, Table~\ref{tab:main_inference} compares the test-time adaptation time of the baseline models and our method applied to TENT~\cite{tent}.
% For the TTA performance, we average the long-term adaptation performance of closed-set and open-set TTA for each dataset.
% The test-time adaptation time in Table~\ref{tab:main_inference} indicates the time consumed for the forward process and the backpropagation.
% Since we utilize the outputs of $\theta_o$, our method accompanies an additional forward process.
% However, as shown, such an additional forward process is negligible compared to the state-of-the-art models such as SWR~\cite{swr} and EATA~\cite{eata}.
% Additionally, we even outperform their TTA performance even with reduced test-time adaptation time. 
% Thus, we want to emphasize that our method brings a significant performance gain with a negligible or smaller amount of time.

\subsection{Comparisons on Resource Costs}
\vspace{-0.1cm}
Along with the TTA performances, Table~\ref{tab:main_inference} compares the memory usage and the time consumption of the baseline models and our method applied to TENT~\cite{tent}.
For the TTA performance, we average the long-term adaptation performance of closed-set and open-set TTA for each dataset.
For memory usage, we use the official code of TinyTL~\cite{tinytl} to calculate both the model parameters and the intermediate activation size, following the previous studies~\cite{back_razor,repnet,ecotta}.
The time indicates the amount of time consumed for the forward process and the backpropagation.
Since we utilize the outputs of $\theta_o$, our method accompanies an additional forward process.
However, as shown, such an additional forward process is negligible compared to the state-of-the-art models.
% For example, our method applied to TENT brings a significant performance gain with only a half amount of memory and time compared to SWR~\cite{swr}.
For example, our method applied to TENT brings a significant performance gain with only half the memory and time compared to SWR~\cite{swr}.
Further details on resource costs, along with the results on semantic segmentation, are included in our supplementary.

\vspace{-0.1cm}
\subsection{Applicability on Various Models}
\vspace{-0.2cm}
Since our method focuses on improving the pseudo-labeling quality of entropy minimization, it does not rely on model architectures. 
Table~\ref{tab:main_model_agnostic} shows that applying our method consistently outperforms the baseline models with ResNet50~\cite{resnet} and WideResNet28~\cite{wrn} that were used in previous TTA studies~\cite{swr,lim2023ttn}.
Such results demonstrate that our method is widely applicable to various architectures.

\vspace{-0.1cm}
\subsection{Robustness to Hyper-parameters}
\vspace{-0.2cm}
In real-world applications, we may not know an adequate learning rate before encountering test samples or may not use an optimal batch size due to memory constraints.
In such a case, we need an approach with a stable performance regardless of such hyper-parameters. 
% Table~\ref{tab:main_hyper_param} shows the robustness of our method to hyper-parameters including learning rates and batch sizes.
% We observe that TENT~\cite{tent} shows significant performance variations as it is highly dependent on the learning rates and batch sizes.
% On the other hand, our method consistently outperforms TENT with minimal performance variances. 
Table~\ref{tab:main_hyper_param} shows that our method is more robust to such hyper-parameters compared to TENT~\cite{tent}, which is highly dependent on them.
Such results demonstrate the scalability of our method when we do not know the optimal hyper-parameters.

% %%%%%%%%%%%%%%%%%%%% Full version of related work %%%%%%%%%%%%%%%%%%%%%%%%%%%%%%
\vspace{-0.1cm}
\section{Related Work}
\vspace{-0.1cm}
% \subsection{Test-Time Adaptation} 
\noindent \textbf{Test-Time Adaptation} 
% There exist various studies that address the domain shift between the train set and the test set such as domain generalization~\cite{zhou2021mixstyle, li2022uncertainty, self_reg, domainbed, gradient_surgery, gradient_matching} and unsupervised domain adaptation (UDA)~\cite{sentry, max_square, shot, video_da, long2016unsupervised, uda_without_source}.
The main differences between TTA studies and other studies addressing domain shifts such as domain generalization~\cite{zhou2021mixstyle, li2022uncertainty, self_reg, domainbed, gradient_surgery, gradient_matching} or unsupervised domain adaptation (UDA)~\cite{sentry, max_square, shot, long2016unsupervised, uda_without_source} is that TTA studies do not utilize 1) the source data during the adaptation stage and 2) ground truth labels on the target distribution~\cite{tent,cotta,eata,conjugate,swr,lim2023ttn}. 
% To improve the adaptation performance in such a setting, TENT only updates the affine parameters using the entropy minimization~\cite{tent}.
% SWR \& NSP further improved the TTA performance by 1) updating domain-sensitive weight parameters more than the insensitive ones and 2) aligning the prototype vectors of the source and the target distributions~\cite{swr}. 
Recent studies~\cite{cotta, lim2023ttn, tta_promt} show that TTA models suffer from a large performance degradation with continually changing domains and a long-term adaptation.
To tackle such a challenging problem, this paper mainly evaluates long-term adaptation with continually changing domains. 

% \subsection{Addressing Noisy Signals in Test-Time Adaptation}
% \subsection{Noisy Signals in Test-Time Adaptation}
\vspace{0.03cm}
\noindent \textbf{Noisy Signals in Test-Time Adaptation} 
As aforementioned, one of the key challenges in TTA is that the model is prone to utilizing wrong predictions.
Preventing the model from learning with noisy supervision has been studied widely beyond TTA~\cite{gce, sentry, mentornet, coteaching, noise_modeling, biasensemble, park2023deep}.
However, the main difference between TTA and these studies is that TTA studies assume that we cannot revisit the sample after performing adaptation with it. 
Such an assumption limits from proposing methods that require knowledge of the full data distributions~\cite{selfie,noise_modeling} or consistency of predictions for a given sample~\cite{sentry, morph}.
% Following are the recent studies that improved the pseudo-labeling performance in TTA without such knowledge.
% Conjugate pseudo labels are proposed based on the observation that conjugate functions are approximate to the optimal loss functions~\cite{conjugate}. 
% To tackle a similar problem, EATA selects high-confident (i.e., low-entropy) samples based on the finding that low-confident samples are mostly noisy ones~\cite{eata}.
% To tackle a similar problem without such knowledge in TTA, EATA selects low-entropy samples based on the finding that high-entropy samples are mostly noisy ones~\cite{eata}.
Without such knowledge, we use the difference of confidence scores between $\theta_o$ and $\theta_a$ by using the wisdom of crowds to improve pseudo labeling. 

\vspace{-0.1cm}
\section{Conclusion}
\vspace{-0.13cm}
This paper proposed a \textit{simple yet effective} data sample selection that is widely applicable to existing various test-time adaptation methods.
% Based on the observation that wrong samples fail to increase the confidence values of its predicted label even with entropy minimization, we only select those achieving higher confidence values on the originally predicted class.
Based on the observation that signals from wrong samples fail to increase the confidence values of the predicted labels even with entropy minimization, we only select the samples that achieve higher confidence values with the adaptation model compared to those with the original model.
This is mainly due to the wisdom of crowds, the dominant signals generally found in the correct samples influencing signals of other samples. 
Our method improved TTA performance on the existing TTA methods on both image classification and semantic segmentation. 
Additionally, we proposed a novel evaluation setting, an open-set TTA, which was overlooked until now even with its importance and practicality.
% We hope our work inspires future researchers to improve TTA performance on both closed-set and open-set settings. 
% We hope our work inspires future researchers to improve both closed-set and open-set TTA. 
We hope our work inspires future researchers to conduct more practical TTA research that improves both closed-set and open-set TTA.

\noindent \textbf{Acknowledgement}
We would like to thank Kyuwoong Hwang, Sungrack Yun, Simyung Chang, Hyunsin Park, Janghoon Cho, Juntae Lee, Hyoungwoo Park, Seokeon Choi, Seunghan Yang, and Sunghyun Park of the Qualcomm AI Research team for their valuable discussions.

{\small
\bibliographystyle{ieee_fullname}
\bibliography{main}
}

\clearpage

\appendix

\begin{figure*}
    \centering
    \includegraphics[width=\textwidth]{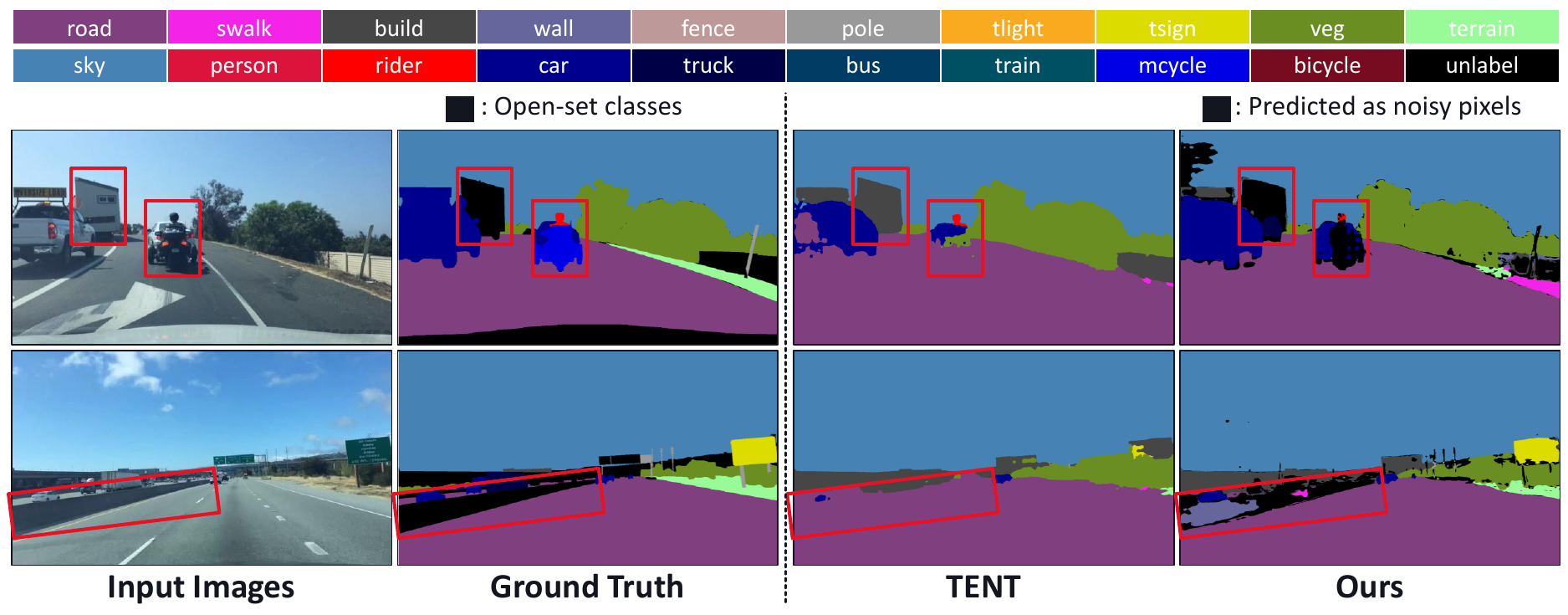}
    \vspace{-0.5cm}
    \captionof{figure}{Identifying wrong predictions and open-set samples in test-time adaptation (TTA). During the long-term adaptation, previous models not only show a large performance degradation but also predict the open-set samples as one of the pre-defined classes learned during the training phase. By filtering out noisy ones, both wrong and open-set samples, we can (a) prevent performance degradation and (b) identify unexpected obstacles to prevent accidents immediately. Red boxes indicate the regions of pixels that include misclassified predictions or open-set classes. In the fourth column, on top of the prediction of the model trained through our method, we color pixels that are filtered out by our method as black.}
    \label{fig:supple_teaser}
\end{figure*}

\section{Further Analysis on Semantic Segmentation}
Fig.~\ref{fig:supple_teaser} shows how filtering out noisy samples is important in semantic segmentation.
As mentioned in the main paper, discarding noisy samples is crucial in two aspects: we can 1) prevent significant performance degradation caused by noisy samples and 2) immediately identify unknown objects that could be highly dangerous if not detected. 
For example, TENT~\cite{tent} predicts the motorcycle (wrong prediction) or the guardrails (open-set predictions) as roads in the first and the second rows, respectively. 
When applying TTA in real-world applications (\eg autonomous driving), such an issue could lead to a serious accident. 
However, our method effectively identifies them immediately (black pixels in the fourth column), which can prevent such accidents. 
%#########################################  previous  ######################################################## 
\begin{table}[b]
\centering
%#################################################
% Rounds 100
%#################################################
\vspace{-0.5cm}
{
\centering

\begin{center}
{\resizebox{0.49\textwidth}{!}{
{

% Method & BDD-100K & Mapillary & GTAV & SYNTHIA & Averag

\begin{tabular}{l|ccc|c} \toprule
Time & \multicolumn{3}{l|}{$t\xrightarrow{\hspace*{7cm}}$} & \multirow{2}{*}{Average} \\ 
\cmidrule(r){0-3}
Method & \multicolumn{1}{c}{Cityscapes} & \multicolumn{1}{c}{BDD-100K} & \multicolumn{1}{c|}{Mapillary} &  \\ \toprule
% Source~\cite{deeplab}                & 34.74 & 16.15 & 36.97 & 29.29 \\
% BN Adapt~\cite{tbn}   & 40.77 & 25.21 & 39.10 & 35.03 \\
% TTN~\cite{lim2023ttn}        & 46.28 & 28.07 & 45.46 & 39.94\\
% \hline
TENT~\cite{tent}                    & 46.73 & 29.59 & 35.69 & 37.34 \\
TENT w/o open-set~\cite{tent}       & \textbf{47.04} & \textbf{31.12} & 38.66 & 38.94 \\
\rowcolor{Gray}+ Ours & 46.76 (+0.03) & 30.55 (+0.96) & \textbf{43.42 (+7.73)} & \textbf{40.24 (+2.90)} \\
\hline
\end{tabular}
}}}
\end{center}
\vspace{-1.em}
}
\caption{Effect of removing open-set samples in semantic segmentation. We filtered out open-set pixels using ground-truth labels for TENT. We observe performance gain compared to the original performance of TENT.}
\label{tab:supple_semantic_segmentation_ood} 
\vspace{-0.15cm}
\end{table}
%##################################################################################################

%%% closed / open version
% %#################################################################################################
% \begin{table}[t]
% \centering
% %#################################################
% % Rounds 50
% %#################################################
% \begin{center}
% {\resizebox{0.45\textwidth}{!}{
% {
% \begin{tabular}{c|cc|c} 
% \toprule
% \multirow{2}{*}{Method} & \multicolumn{2}{c|}{ImageNet-C} & \multirow{2}{*}{Average}   \\
%        & Closed & Open \\ 
% \toprule
% Source~\cite{wrn} & 18.27 & 18.27 \\
% BN Adapt~\cite{tbn} & 14.49 & 15.73 \\
% \hline
% TENT~\cite{tent} & 45.84 & 85.22\\
% \rowcolor{Gray}+ Ours & \textbf{14.10 (-31.74)} & \textbf{15.77 (-69.45)} \\
% \hline
% SWR~\cite{swr} & 10.21 & 90.55 \\
% \rowcolor{Gray}+ Ours & \textbf{10.12 (-0.09)} & \textbf{72.58 (-17.97)} \\
% \toprule
% \end{tabular}}}}
% \end{center}
% \vspace{-0.6cm}
% \caption{Error rates of image classification on ImageNet-C after 10 rounds of adaptation (\ie long-term test-time adaptation). We note the performance gain by reduced error rates.}
% \label{tab:supple_imagenetc} 
% \vspace{-0.5cm}
% \end{table}
%##################################################################################################

\begin{table*}[t]
\centering
%#################################################
% Rounds 100
%#################################################
\subfloat[Error rates after 10 rounds of adaptation.\label{tab:supple_imagenetc_long}]{
\centering
\begin{minipage}{0.48\linewidth}{
\begin{center}
{\resizebox{\textwidth}{!}{
{
\begin{tabular}{c|cc|c} 
\toprule
\multirow{2}{*}{Method} & \multicolumn{2}{c|}{ImageNet-C} & \multirow{2}{*}{Average}   \\
       & Closed & Open \\ 
\toprule
Source~\cite{wrn} & 81.99 & 81.99 & 81.99 \\
BN Adapt~\cite{tbn} & 68.49 & 69.65 & 69.07 \\
\hline
TENT~\cite{tent} & 99.71 & 99.72 & 99.72 \\
\rowcolor{Gray}+ Ours & \textbf{65.62 (-34.09)} & \textbf{67.78 (-31.94)} & \textbf{66.70 (-33.02)}  \\
\hline
SWR~\cite{swr} & 65.20 & 68.40 & 66.80 \\
\rowcolor{Gray}+ Ours & \textbf{64.35 (-0.85)} & \textbf{66.33 (-2.07)} & \textbf{65.34 (-1.46)} \\
\toprule
\end{tabular}}}}
\end{center}
\vspace{-1.5em}
\label{tab:main_cifar} 
}
\end{minipage}
}
%#################################################
% Round 1
%#################################################
\subfloat[Error rates after 1 round of adaptation.\label{tab:supple_imagenetc_short}]{
\centering
\begin{minipage}{0.48\linewidth}{
\begin{center}
{\resizebox{\textwidth}{!}{
{
\begin{tabular}{c|cc|c} 
\toprule
\multirow{2}{*}{Method} & \multicolumn{2}{c|}{ImageNet-C} & \multirow{2}{*}{Average}   \\
       & Closed & Open \\ 
\toprule
Source~\cite{wrn} & 81.99 & 81.99 & 81.99 \\
BN Adapt~\cite{tbn} & 68.49 & 69.65 & 69.07 \\
\hline
TENT~\cite{tent} & 95.79 & 97.53 & 96.66 \\
\rowcolor{Gray}+ Ours & \textbf{60.82 (-34.97)} & \textbf{64.33 (-33.20)} & \textbf{62.58 (-34.08)}  \\
\hline
SWR~\cite{swr} & 66.59 & 69.02 & 67.81 \\
\rowcolor{Gray}+ Ours & \textbf{65.29 (-1.30)} & \textbf{66.86 (-2.16)} & \textbf{66.08 (-1.73)} \\
\toprule
\end{tabular}}}}
\end{center}
\vspace{-1.5em}
}
\end{minipage}
}
\vspace{-0.26cm}
\caption{Comparisons on ImageNet-C. We note the performance gain by reduced error rates.}
\label{tab:supple_imagenetc}
\vspace{-0.1cm}
\end{table*}
\begin{table*}[t]
\centering
%#################################################
% Rounds 100
%#################################################
\subfloat[Image classification\label{tab:supple_cotta_image}]{
\centering
\begin{minipage}{0.48\linewidth}{
\begin{center}
{\resizebox{\textwidth}{!}{
{

\begin{tabular}{c|ccc|ccc|cc} 
\toprule
\multirow{2}{*}{Method} & \multicolumn{3}{c|}{CIFAR-10-C}  & \multicolumn{3}{c|}{CIFAR-100-C} & \multirow{2}{*}{\makecell{Memory \\ (MB)}} & \multirow{2}{*}{\makecell{Time \\ (ms)}} \\
       & (a) & (b) & (c) & (a) & (b) & (c) & (MB) & (ms) \\ 
\toprule
TENT~\cite{tent}    & 56.00 & 45.84 & 45.84 & 45.20 & 42.34 & 42.34 & 556 & 18.38 \\
CoTTA~\cite{cotta}   & 31.28 & 75.97 & 83.19 & 41.40 & 94.52 & 97.43 & 36442 & 379.49 \\
\rowcolor{Gray}TENT + Ours                & \textbf{20.16} & \textbf{14.10} & \textbf{14.10} & \textbf{33.39} & \textbf{38.62} & \textbf{38.62} & 565 & 26.62 \\
\toprule
\end{tabular}}}}
\end{center}
\vspace{-1.5em}
\label{tab:main_cifar}
}
\end{minipage}
}
%#################################################
% Round 1
%#################################################
\subfloat[Semantic Segmentation\label{tab:supple_cotta_segment}]{
\centering
\begin{minipage}{0.49\linewidth}{
\begin{center}
{\resizebox{\textwidth}{!}{
{

\begin{tabular}{l|ccc|c} \toprule
Time & \multicolumn{3}{l|}{$t\xrightarrow{\hspace*{7cm}}$} & \multirow{2}{*}{Average} \\ 
\cmidrule(r){0-3}
Method & \multicolumn{1}{c}{Cityscapes} & \multicolumn{1}{c}{BDD-100K} & \multicolumn{1}{c|}{Mapillary} &  \\ 
\toprule
% TTN~\cite{lim2023ttn}        & 46.28 & 28.07 & 45.46 & 39.94\\
% \hline
TENT~\cite{tent}      & 46.73 & 29.59 & 35.69 & 37.34 \\
CoTTA~\cite{cotta}    & 41.03 & 26.42 & 40.03 & 33.23 \\
\rowcolor{Gray}TENT+ Ours & \textbf{46.76 (+0.03)} & \textbf{30.55 (+0.96)} & \textbf{43.42 (+7.73)} & \textbf{40.24 (+2.90)} \\ \hline
\end{tabular}
}}}
\end{center}
\vspace{-1.5em}
}
\end{minipage}
}
\vspace{-0.26cm}
\caption{Comparison between our method and CoTTA~\cite{cotta}. We show the results of our method applied to TENT. We perform better than CoTTA even with a substantially smaller amount of memory usage and time consumption.}
% We use Tiny-ImageNet-C for the evaluation. Our method shows a stable performance with low standard deviations compared to TENT~\cite{tent}.
\label{tab:supple_cotta}
\vspace{-0.2cm}
\end{table*}
%##################################################################################################

Table~\ref{tab:supple_semantic_segmentation_ood} shows that open-set samples degrade the performance of TTA models in semantic segmentation. 
% Along with the results reported in Table~\ref{tab:supple_semantic_segmentation_ood}, we also report the result of TENT~\cite{tent} trained without the backpropagation of the open-set pixels.
For the analysis, we compare the performance of TENT~\cite{tent} and that of TENT trained without the backpropagation of the open-set pixels. 
We use the ground truth labels and filter out the open-set pixels.
As shown, TENT achieves better performance without the backpropagation of the open-set pixels compared to the original performance.
Such a result again demonstrates that addressing open-set samples is crucial for practical TTA. 
Note that our approach still outperforms TENT adapted with open-set samples filtered out after a long-term adaptation (\eg Mapillary).
This is mainly due to the fact that our method discards the wrong predictions well in addition to the open-set samples. 

% \vspace{-0.1cm}
\section{Comparisons on ImageNet-C}
In Table~\ref{tab:supple_imagenetc}, we also verify the effectiveness of our method on a large-scale dataset, ImageNet-C~\cite{imagenetC}.
Due to the fact that experimentation on ImageNet-C is time consuming, we simulate the long-term adaptation with 10 rounds instead of the 50 rounds used in the main paper. 
We evaluate under continuously changing target domains without resetting the model between each domain.
We use the batch size of 64 and the learning rate of 0.00025 with the SGD optimizer~\cite{sgd}, following the previous studies~\cite{tent,lim2023ttn,swr}.
We observe that our method again consistently improves the TTA performance on existing baseline models in closed-set and open-set settings with short-term and long-term adaptation.
Regarding SWR~\cite{swr}, we observe a significant performance drop of SWR when utilizing the adapted model of the previous iteration for the regularization. 
Therefore, we use the source pretrained original model, $\theta_o$, for the regularization. 
Other hyper-parameters are set as the default values.

\section{Comparisons with CoTTA}
We also compare our method with CoTTA~\cite{cotta}, another seminal work in the continual test-time adaptation. 
Table~\ref{tab:supple_cotta} compares the performances of image classification and semantic segmentation and the resource costs between CoTTA and our method applied to TENT~\cite{tent}.
As shown, although our method utilizes a significantly smaller amount of memory usage and time consumption, we achieve better performance in both image classification and semantic segmentation. 
We describe the results in detail.

\begin{table}[b]
%#################################################
% Rounds 100
%#################################################
\vspace{-0.15cm}
{\centering

\begin{center}
{\resizebox{0.4\textwidth}{!}{
{
\begin{tabular}{c|cc} 
\toprule
Method & Memory (MB) & Time (ms) \\
\hline
% ENT                         & 5815 & xxx \\ 
TENT~\cite{tent}            & 2714 & 529 \\ 
SWR~\cite{swr}              & 5969 & 625 \\ 
CoTTA~\cite{cotta}          & 20276 & 4499 \\ 
\rowcolor{Gray} TENT~\cite{tent} + Ours       & 3036 & 685 \\ 

\hline
\toprule
\end{tabular}}}}
\end{center}
\vspace{-0.35cm}
}
\caption{Comparisons on memory usage (MB), and time consumption (ms) on semantic segmentation. We evaluate with DeepLabV3Plus-ResNet-50~\cite{deeplab}. For memory usage, we use the batch size of 2. For the time, we report the average time after 5000 trials with the image resolution of 3$\times$800$\times$1455 on NVIDIA RTX A5000.}
\vspace{-0.15cm}
\label{tab:supple_time} 
\end{table}

\subsection{Image Classification}
% Regarding image classification, we use the model architectures reported in the main paper of CoTTA (\ie WideResnet28~\cite{wrn} for CIFAR-10-C and ResNeXt-29 for CIFAR-100-C) since 
We observe that CoTTA~\cite{cotta} shows performance variations depending on the hyper-parameter $p_{th}$, which is a threshold to decide whether to use ensembled predictions or a single prediction in CoTTA.
However, we found it challenging to find adequate $p_{th}$ for CoTTA with the model architecture used in our work (\ie WideResNet40~\cite{wrn} for both CIFAR-10-C and CIFAR-100-C). 
Although the supplementary of CoTTA illustrates how to find $p_{th}$, we could not obtain identical values by using the architecture used in CoTTA even with the description.
Therefore, we report the comparisons between CoTTA and our method with the following experimental setups: a) architectures used in the CoTTA paper (\ie WideResNet28~\cite{wrn} for CIFAR-10-C and ResNeXt-29 for CIFAR-100-C) with their default $p_{th}$ values, b) architectures used in our main paper with their default $p_{th}$ values, c) architectures used in our main paper with $p_{th}$ values we found by following the description of the supplementary of CoTTA.
Table~\ref{tab:supple_cotta_image} shows that our method outperforms CoTTA in all three cases even with a substantially smaller amount of memory usage and time consumption. 
For the experiments, we use the official repository of CoTTA\footnote{\scriptsize \href{https://github.com/qinenergy/cotta}{https://github.com/qinenergy/cotta}}.

\subsection{Semantic Segmentation}
Regarding semantic segmentation, we evaluate CoTTA with continuously changing target domains with a model pretrained on GTAV, as done in the main paper. 
% Regarding the multi-scaling utilized in CoTTA, we use the multi-scale factors of [0.5, 0.75, 1.0, 1.25, 1.5] and flip.
While TENT~\cite{tent} and our method show performance gains by using TTN~\cite{lim2023ttn}, CoTTA achieves better performance by utilizing batch normalization with the test statistics (\ie TBN) than by using TTN. 
Therefore, we report the performance of CoTTA using the TBN and the results of TENT and ours using TTN.
In Table~\ref{tab:supple_cotta_segment}, we again observe that our method outperforms CoTTA with real-domain shifts in semantic segmentation.

Additionally, we compare the memory usage and time consumption of our method applied to TENT and other baseline models on semantic segmentation in Table~\ref{tab:supple_time}.
As shown, our method accompanies a negligible amount of resource cost.
For example, while our method outperforms CoTTA, we accompany a substantially smaller amount of resource cost compared to CoTTA.

\section{Further Details on Experimental Setup}
% We further describe the experimental setup in addition to Section~3.1.
\subsection{Datasets}
\begin{figure}
    \centering
    \includegraphics[width=0.49\textwidth]{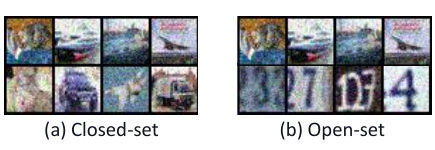}
    \vspace{-0.6cm}
    \captionof{figure}{Examples of datasets used in our work. We use CIFAR-10-C and SVHN-C for the images. In the closet-set TTA, all images in the mini-batch only include the covariate shift (\ie domain shift). On the other hand, in the open-set TTA, half of the images in the mini-batch only include covariate shift while the other half includes both covariate shift and semantic shift (\ie open-set samples).}
    \vspace{-0.5cm}
    \label{fig:supple_dataset_description}
\end{figure}

\noindent \textbf{Image classification}
For constructing SVHN-C and Imagenet-O-C, we apply corruption types used for CIFAR-10/100-C and TinyImagnet-C by using the official code\footnote{\scriptsize \href{https://github.com/hendrycks/robustness}{https://github.com/hendrycks/robustness}} of Hendrycks~\cite{imagenetC}.
Since the image sizes of Imagenet-O~\cite{imagenet_o} and TinyImageNet~\cite{tinyimagenet} are different, we resize the resolution of Imagenet-O images to 64$\times$64.
Among the 5 severity levels, we use corruption level 5, the most corrupted version.
Fig.~\ref{fig:supple_dataset_description} shows the example images of the datasets used in our work.

\noindent \textbf{Semantic segmentation}
For the experiments with continuously changing domains, we use the train sets of each target domain in order to conduct experiments with a long-term adaptation without using multiple rounds. 
Note that each target domain includes a different number of images.
For example, Cityscapes, BDD-100K, and Mapillary include 2975, 7000, and 18000 images, respectively. 
Due to this fact, for showing the mIoU changes in Table~3b of the main paper, we evaluate models an equal number of times (\ie 20 times) for each target domain, not after certain steps. 
For the experiment with a fixed target domain over multiple rounds, we use the validation sets of each target domain.

% \subsection{Evaluation Settings}
\subsection{Baselines}
\noindent \textbf{Conjugate}~\cite{conjugate}
Conjugate pseudo labeling was recently proposed on the observation that conjugate functions are approximate to the optimal loss function. 
We use the official codes\footnote{\scriptsize \href{https://github.com/locuslab/tta_conjugate}{https://github.com/locuslab/tta\_conjugate}} of Conjugate~\cite{conjugate}.

\noindent \textbf{GCE~\cite{gce}}
Generalized Cross Entropy (GCE) loss was first introduced to address the noisy labels in image classification. 
It emphasizes the learning of correct samples by imposing high weights on the gradients of the samples achieving low loss values, which are highly likely to be correctly annotated.
Following Conjugate~\cite{conjugate}, we use GCE as the baseline model to show that simply applying existing noisy-labeling studies does not guarantee preventing the error accumulation in TTA. 
Since the official repository of Conjugate includes GCE codes, noted as RPL, we use the same codes in our work. 

\noindent \textbf{EATA}~\cite{eata}
EATA\footnote{\scriptsize \href{https://github.com/mr-eggplant/EATA}{https://github.com/mr-eggplant/EATA}} filters out samples that achieve loss values higher than a pre-defined static threshold and utilizes the fisher regularization to prevent catastrophic forgetting. 
For the fisher regularization, the original paper utilizes the \textit{test set} of the source distribution to obtain the weight importance $w(\theta)$. 
However, we believe that such an assumption is not valid since the currently widely used corrupted test sets apply the corruptions to the test samples of the source distribution.
In other words, such an approach necessitates the test samples to obtain the weight importance before encountering the test samples. 
Therefore, we use the \textit{train set} of the source distribution to obtain the weight importance. 
For the fisher coefficient, we use 1 for CIFAR-10/100-C and 2000 for TinyImageNet-C, which are the default values reported in the main paper.
For applying our method to EATA, we only replace the filtering method and utilize the fisher regularization. 

\noindent \textbf{SWR~\cite{swr}}
SWR proposes 1) updating domain-sensitive weight parameters more than the insensitive ones and 2) aligning the prototype vectors of the source and the target distributions~\cite{swr}. 
Since SWR does not have an official repository, we re-implemented the codes and report the results. 

\subsection{Implementation Details}
\noindent \textbf{Image classification}
For the image classification on CIFAR-10/100-C, we mainly use WideResNet40~\cite{wrn} which applied the AugMix~\cite{augmix} during the pre-training stage, following the previous recent TTA studies~\cite{lim2023ttn, swr, ecotta}.
The pretrained model is available from RobustBench~\cite{robustbench}.
For the TinyImageNet-C, we use ResNet50~\cite{resnet}.
We pretrained ResNet50 for 50 epochs with a batch size of 256 and a learning rate of 0.01 with cosine annealing applied using the SGD optimizer~\cite{sgd}. 
% We set $\lambda_{max}=0.5$ for experiments on SWR and $\lambda_{max}=0.25$ for the rest of the experiments. 
We set $\lambda_{max}=0.5$ for all experiments. 

\noindent \textbf{Semantic segmentation}
For all semantic segmentation experiments which utilize the backpropagation, we use TTN~\cite{lim2023ttn} since it brings further performance gain compared to using TBN.
For applying our method on semantic segmentation, we use a relaxed version: we select pixels achieving $\hat{y}^{c_o} - \Tilde{y}^{c_o} \geq -0.2$. 
For applying our method on SWR, we reduce the coefficient of the mean entropy maximization loss ($\lambda_{max}$) from 0.5 to 0.2.
The main reason is that the mean entropy maximization works as regularization and reduces the effect of entropy minimization loss.
However, since our work improves the quality of entropy minimization, the mean entropy maximization rather hampers further performance gain from our method. 
By reducing the coefficient of mean entropy maximization, our method improves the semantic segmentation performance.
Such an observation again demonstrates that our method improves the quality of the entropy minimization loss. 
We set other hyper-parameters as the default values.

\vspace{-0.2cm}
\section{Further Details on Resource Costs}
We illustrate how we compute the resource costs including memory usage and time consumption.
For memory usage, as mentioned in the main paper, we use the official code provided by TinyTL~\cite{tinytl}.
Note that the activation size occupies memory usage more than the parameter size~\cite{tinytl,ecotta}. 
For ENT, which updates all parameters, we add the parameter size and activation size of all parameters.
For TENT~\cite{tent}, which updates the affine parameters in the batch normalization layers, we only add the parameter size and activation size of the affine parameters.
For SWR~\cite{swr}, which updates all parameters and utilizes an additional model for the regularization, we add the parameter size of the whole model parameters in addition to the memory usage of ENT.
For EATA~\cite{eata}, which also utilizes an additional model for the fisher regularization, we only add the parameter size of the affine parameters in addition to the memory usage of TENT.
For our method applied to TENT, in addition to the memory usage of TENT, we add 1) the parameter size of all parameters and 2) the parameter size of the output tensors.
We add the parameter size of all parameters since we need the whole model parameters in order to compute $\Tilde{y}$.
Also, since we utilize $\Tilde{y}$, we add the memory of the output tensors that is negligible compared to the parameter size of the whole model.

\begin{table}[t]
\centering
%#################################################
% Rounds 50
%#################################################
\begin{center}
{\resizebox{0.49\textwidth}{!}{
{
\begin{tabular}{c|ccc} 
\toprule
Method & CIFAR-10-C  & CIFAR-100-C & TinyImageNet-C \\
\toprule
Source~\cite{wrn} & 18.27 & 46.75 & 76.71 \\
BN Adapt~\cite{tbn} & 14.49 & 39.26 & 61.90 \\
\hline
TENT~\cite{tent} & 45.84 & 42.34 & 98.10 \\
\rowcolor{Gray}+ Ours (logit) & 33.46 (-12.38) & 72.08 (+29.74) & 92.24 (-5.86) \\
\rowcolor{Gray}+ Ours (softmax) & \textbf{14.10 (-31.74)} & \textbf{38.62 (-3.72)} & \textbf{60.87 (-37.23)} \\
\toprule
\end{tabular}}}}
\end{center}
\vspace{-0.5cm}
\caption{Variant of our method. We observe that utilizing the softmax outputs outperforms utilizing the logit values.}
\label{tab:supple_variant} 
\vspace{-0.2cm}
\end{table}
%##################################################################################################

\section{Variant of Proposed Method}
To compare the prediction values between $\theta_a$ and $\theta_o$, our method utilizes the probability values of the softmax outputs. 
In Table~\ref{tab:supple_variant}, we also analyze our method by using the logit values instead of the softmax values. 
We observe that utilizing logit values fails to bring large performance gains compared to using the softmax values. 
The main reason is that the logit values generally increase regardless of the correct or wrong samples. 
However, such an issue is not found in the softmax outputs since the values are normalized to sum-to-one vectors.

\end{document}